\documentclass[letterpaper, 10 pt, conference]{ieeeconf}  % Comment this line out if you need a4paper

\IEEEoverridecommandlockouts                              % This command is only needed if 
\usepackage{graphicx}
\usepackage{cite}
\usepackage{amsmath,amssymb,amsfonts}
\usepackage{algorithmic}
\usepackage{textcomp}
\usepackage{subcaption}
\usepackage{caption}
\usepackage{booktabs} 
 \usepackage{multirow}

\title{\LARGE \bf
LiV-GS: LiDAR-Vision Integration for 3D Gaussian Splatting SLAM in Outdoor Environments
}

\author{Renxiang Xiao$^{\dagger}$, Wei Liu$^{\dagger}$, Yushuai Chen, Liang Hu*% <-this % stops a space
\thanks{$^{\dagger}$ Equal Contribution, * For correspondence (l.hu@hit.edu.cn)}
\thanks{R. Xiao, W. Liu, Y. Chen and L. Hu are with the Department of Automation, School of Mechanical Engineering and Automation, Harbin Institute of Technology, Shenzhen, China. }%
}

\begin{document}

\maketitle
\thispagestyle{empty}
\pagestyle{empty}

%%%%%%%%%%%%%%%%%%%%%%%%%%%%%%%%%%%%%%%%%%%%%%%%%%%%%%%%%%%%%%%%%%%%%%%%%%%%%%%%
\begin{abstract}
We present LiV-GS, a LiDAR-visual SLAM system in outdoor environments that leverages 3D Gaussian as a differentiable spatial representation. Notably, LiV-GS is the first method that directly aligns discrete and sparse LiDAR data with continuous differentiable Gaussian maps in large-scale outdoor scenes, overcoming the limitation of fixed resolution in traditional LiDAR mapping. The system aligns point clouds with Gaussian maps using shared covariance attributes for front-end tracking and integrates the normal orientation into the loss function to refines the Gaussian map. To reliably and stably update Gaussians outside the LiDAR field of view, we introduce a novel conditional Gaussian constraint that aligns these Gaussians closely with the nearest reliable ones. The targeted adjustment enables LiV-GS to achieve fast and accurate mapping with novel view synthesis at a rate of 7.98 FPS. Extensive comparative experiments demonstrate LiV-GS's superior performance in SLAM, image rendering and mapping. The successful cross-modal radar-LiDAR localization highlights the potential of LiV-GS for applications in cross-modal semantic positioning and object segmentation with Gaussian maps.

\end{abstract}

%%%%%%%%%%%%%%%%%%%%%%%%%%%%%%%%%%%%%%%%%%%%%%%%%%%%%%%%%%%%%%%%%%%%%%%%%%%%%%%%
\section{INTRODUCTION}

SLAM (Simultaneous Localization and Mapping) is essential for large-scale scene reconstruction which rebuilds observed scenes based on precise positioning. As two commonly used sensors for scene reconstruction, LiDAR and cameras have shaped the development of SLAM technology. Traditional LiDAR SLAM exploits geometric and precise depth information in point clouds for accurate positioning. The recent advent of neural radiance fields (NeRFs)\cite{mildenhall2021nerf} and 3D Gaussian Splatting (3DGS)\cite{kerbl20233dgs} has enriched visual SLAM with more detailed environmental representations. Traditional map representations such as point clouds, voxels, and surfels suffer from bounded-resolution in mapping\cite{deng2023nerfloam}, while implicit neural fields demand expensive computational resources. In contrast, 3DGS uses Gaussian ellipsoids as a highly effective map, allowing for adaptive spatial feature representation and fast rendering. Considering the needs of accurate SLAM and photo-realistic scene reconstruction, 3D Gaussian splatting emerges as a powerful approach that effectively combines the advantages of both LiDAR and camera sensors.

Outdoor SLAM and scene reconstruction face unique challenges, such as lighting variations and unbounded depth scales, which make indoor RGBD-based solutions inadequate \cite{ha2024rgbdgsicpslam,tao2024silvr,yan2024gsslam,sun2024mm3dgs,islam2024mvslam}. While LiDAR excels in precise distance measurement in outdoor environments, maintaining pixel correspondence across different views is challenging due to the sparse and discontinuous nature of LiDAR point clouds. Existing methods such as handheld mapping, rely on repetitive mapping to enhance image depth, which is impractical for mobile robot platforms that capture only a single view from the camera's perspective. 

To the end, we introduce LiV-GS, a SLAM framework that uses 3D Gaussian spatial representations to seamlessly integrate LiDAR and camera images. Our method estimates robot pose by aligning Gaussian covariance from rendering with the current observations, with the back-end correcting drift and updating the Gaussian map. To overcome depth continuity issues between vision and LiDAR in unbounded scenes, we propose a Gaussian splitting method based on LiDAR point clouds, ensuring proper distribution constraints in map updates.

Our research contributions are summarized as follows:
\begin{enumerate}
\item We propose a unified LiDAR-camera outdoor SLAM framework using 3D Gaussian representations, enabling incremental mapping and high-quality new view synthesis during the process of high-precision positioning;
\item We introduce effective Gaussian-LiDAR alignment methods, including a normal direction constraint for stable tracking and a density- and normal-consistency-based weighting mechanism to account for the reliability of different Gaussians;
\item We propose a conditional Gaussian distribution constraint for map updates, allowing the propagation of reliable Gaussians with LiDAR priors to represent the entire scene, even including objects or segments where LiDAR points are unavailable.
\end{enumerate}

\begin{figure*}[ht]
    \centering
    \includegraphics[width=0.9\textwidth]{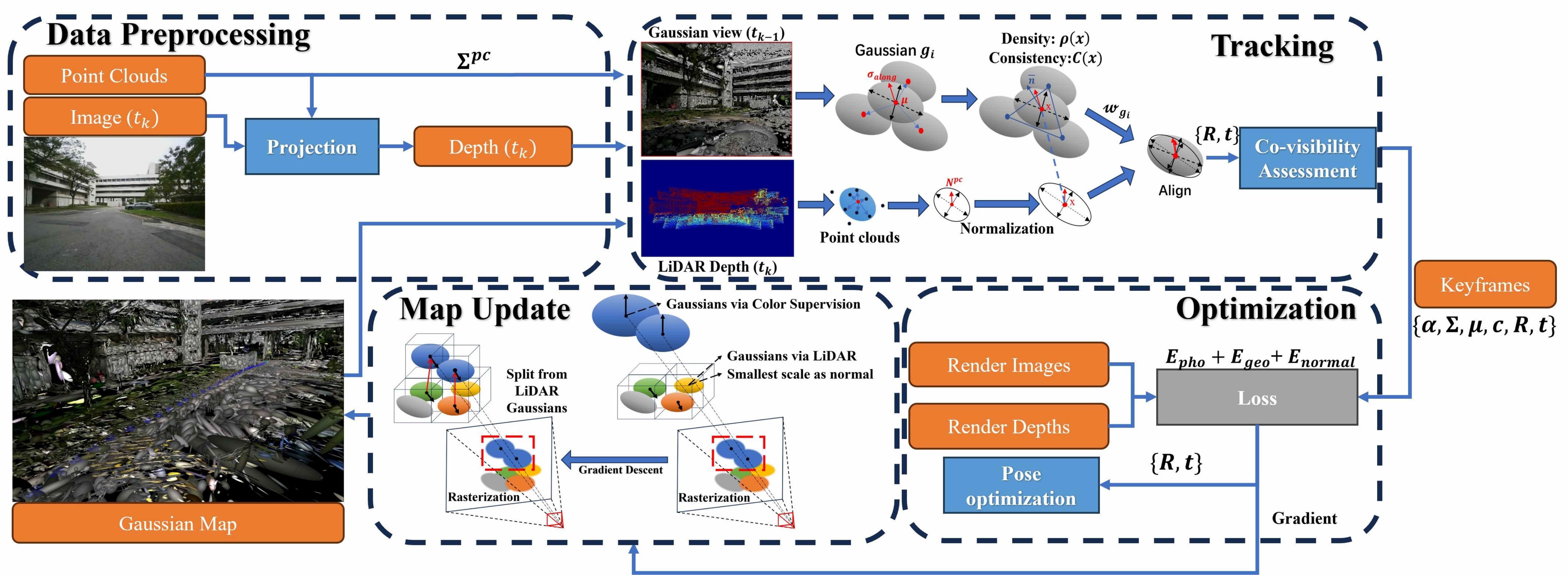}
    \caption{\textbf{Overview of the system:} The SLAM system comprises a tracking and optimization process that together support the visual representation of the Gaussian map. The map update process uses LiDAR depth and color supervision to adjust the new Gaussians.}
    \label{flowchart}
\vspace{-3mm}
\end{figure*}

\section{Related Works}

\subsection{Large Scale Reconstruction}
Existing tasks for large-scale scene reconstruction rely on known poses and sparse point clouds to render environments. For unbounded large scenes, major improvements have primarily focused on constraining Gaussian map from a Bird's Eye View (BEV) perspective, leveraging Levels of Detail (LoD), or partitioning rendering regions to accelerate rendering and enhance computational stability. Extending on the original work of 3D Gaussian Splatting, \cite{kerbl2024hierarchical} proposes a tree-like structure that uses the projected size within a view to selectively include or exclude specific nodes, enhancing rendering efficiency. The research work DoGaussian\cite{chen2024dogaussian} introduces improvements through partitioned rendering techniques. GaussianCity in \cite{xie2024gaussiancity} innovatively handles point clouds by transforming the visual perspective into BEV and employing an encode-decode strategy for efficient processing. Octree-GS\cite{ren2024octree-gs} utilizes the octree data structure to effectively manage multi-level details of a scene. EfficientGS\cite{liu2024efficientgs} incorporates a selective densification strategy and applies sparse order increments in spherical harmonics (SH). GaussianPro\cite{cheng2024gaussianpro} introduces a progressive propagation approach for Gaussians, using depth and normal constraints to render large textureless scene areas effectively.

\subsection{3DGS-based LiDAR-Visual Fusion}
The integration of LiDAR data significantly enhances the capabilities of 3DGS in large-scale scene reconstruction. The utilization of LiDAR involves using point clouds maps and poses obtained from LiDAR SLAM as initial inputs, replacing traditional SfM data. For instance, DrivingGaussian\cite{zhou2024drivinggaussian} extracts image features and projects merged LiDAR scan frames, enhancing feature integration. 3DGS-ReLoc\cite{jiang20243dgsReloc} and Gaussian-LIC\cite{lang2024gaussianLIC} train 3D maps using direct inputs from LiDAR point clouds and corresponding images, achieving more detailed environmental models. LIV-GaussianMap\cite{hong2024livGaussianmap} advances point clouds management by employing an explicit octree structure, while TCLC-GS\cite{zhao2024tclcgs} builds a hierarchical octree of implicit feature grids, using color projection for further 3DGS optimization. LetsGo\cite{cui2024letsgo} integrates Level of Detail (LoD) rendering into 3DGS, using Gaussian functions at various resolutions to represent 3D scenes efficiently. 

Another approach integrates 3DGS directly into the SLAM process. For example, MM-Gaussian \cite{wu2024mmgaussian} uses a point cloud registration algorithm to estimate camera pose and directly merges these point clouds into the map for enhanced optimization. This method enables visualization of incremental mapping and leverages high-fidelity scene reconstruction to reverse-optimize pose, achieving near-optimal positioning accuracy and mapping quality simultaneously. These methodologies collectively highlight the pivotal role of LiDAR in refining 3D reconstruction processes.

The most closely-related works to our research include DrivingGaussian  \cite{zhou2024drivinggaussian}, LIV-GaussMap \cite{ hong2024livGaussianmap}, and MM-Gaussian \cite{wu2024mmgaussian}. As summarized in Table \ref{Lidar-3DGS comparison}, the existing methods typically separate the tasks of SLAM and dense Gaussian mapping, where either LiDAR point clouds are used to replace SfM or point clouds are directly matched as tracking priors, and then the pose is optimized with a rendering loss based on the Gaussian map. In contrast, our LiV-GS framework optimizes both pose estimation and map updates within an integrated Gaussian map representation. Moreover, while existing 3DGS-based outdoor SLAM methods focus solely on regions where both LiDAR point clouds and visual pixels are captured, our approach reconstructs scenes beyond the LiDAR field of view with high quality, thanks to our novel Gaussian splitting method derived from LiDAR-based Gaussians.
\begin{table}[h]
\centering
\caption{\footnotesize{Comparison of 3DGS-based LiDAR-Visual Fusion Methods}}
\label{Lidar-3DGS comparison}\resizebox{0.5\textwidth}{!}{
\begin{tabular}{c|c|c|c}
\hline
\multirow{2}{*}{Methods} & \multirow{2}{*}{Task} & \multirow{2}{*}{Pose acquisition} & \multirow{2}{*}{Region of Scene Reconstruction} \\
        &      &                 & \\
\hline
\multirow{2}{*}{DrivingGaussian \cite{zhou2024drivinggaussian}} & \multirow{2}{*}{Reconstruction} & \multirow{2}{*}{Colmap} & \multirow{2}{*}{Within FOV of Camera and LiDAR range}\\
                &                &          & \\
\hline
\multirow{2}{*}{LIV-GaussianMap \cite{hong2024livGaussianmap}} & \multirow{2}{*}{Reconstruction} & \multirow{2}{*}{LiDAR-inertial SLAM} & \multirow{2}{*}{Within FOV of Camera and LiDAR range}\\
                &                &          & \\
\hline
\multirow{2}{*}{MMGaussian \cite{wu2024mmgaussian}} &Reconstruction & \multirow{2}{*}{Point cloud Match Prior} &\multirow{2}{*}{Within FOV of Camera and LiDAR range}\\
                &     + SLAM           &           & \\
\hline
\multirow{2}{*}{Our LiV-GS} &Reconstruction & LiDAR Point clouds and & \multirow{2}{*}{Within FOV of Camera}\\
     &    + SLAM      & Gaussian map  alignment       & \\
\hline
\end{tabular}
}
\end{table}

% \begin{table*}[h]
% \centering
% \caption{Comparison of LiDAR-3DGS Methods}
% \label{Lidar-3DGS comparison}
% \begin{tabular}{|c|c|c|c|}
% \hline
% Method & Task & Pose acquisition & Ability to reconstruct areas without depth information\\
% \hline
% LIV-GaussianMap & Reconstruction & LiDAR-inertial SLAM & No\\
% \hline
% MMGaussian &Reconstruction+SLAM & Point cloud Match & No\\
% \hline
% Ours &Reconstruction+SLAM & Align LiDAR Point cloud and Gaussian map & Yes\\
% \hline
% \end{tabular}
% \end{table*}

\section{Methodology}
% \begin{figure*}[t]
%     \centering
%     \includegraphics[width=\textwidth]{figures/flowchart.png}
%     \caption{Flowchart of the system}
%     \label{flowchart}
% \end{figure*}

Our LiV-GS is an outdoor visual-LiDAR SLAM system that employs 3D gaussian for environmental representation. The Gaussian distribution is mathematically expressed as:
\begin{equation}
G(x) = e^{-\frac{1}{2}(x-\mu)^{\top} \Sigma^{-1} (x-\mu)}
\end{equation}
where \(\mu\) and \(\Sigma\) denote the mean and covariance matrix of the Gaussians, respectively. In our model, each Gaussian is defined by \({g}_i = \{\alpha, c, \mu, \Sigma\}\), where \(\alpha\) signifies opacity, and \(c\) the color, directly derived from the original pixel data. The same as \cite{matsuki2024gaussianslams}, we omit the spherical harmonics for simplicity and speed.

% Through extensive experiments and observations, we have found that points closer to the object surface are geometrically more stable and exhibit minimal color variation. Thus, we have adopted an innovative approach by substituting the traditional Gaussian covariance with the modified covariance derived from LiDAR point clouds. This substitution introduces new geometric constraints to the Gaussian Gaussians, effectively supervising the splitting of Gaussians during the gradient descent process. Unlike traditional approaches that rely solely on LiDAR point clouds for map construction and visual inputs for pose estimation, each module of our system interacts with the 3D Gaussian map, enhancing both localization accuracy and mapping fidelity.
\begin{figure}[h]
    \centering
    \includegraphics[width=0.9\linewidth]{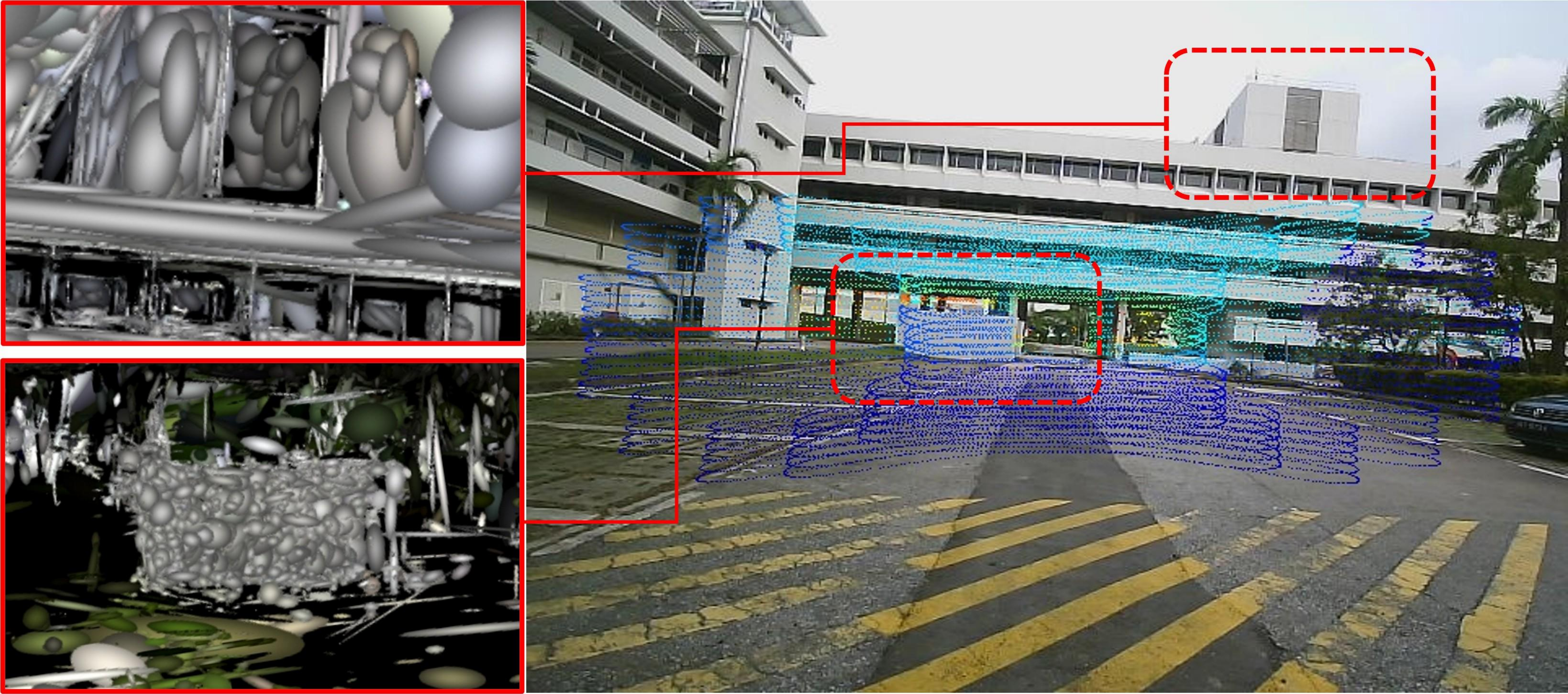}
    \caption{\textbf{Relationship between Density and weight:} Gaussians based on only color supervision result in isotropic and sparse Gaussians (top left). Regions with dense depth input from LiDAR typically show Gaussians in higher density (bottom left).}
    \label{dense}
    \vspace{-5mm}
\end{figure}

\subsection{System overview}
As depicted in Fig. \ref{flowchart}, the entire system of LiV-GS comprises four modules: Data Preporessing, Tracking, Optimization and Map Update. The multi-modal measurements from LiDAR and visual sensors are integrated in Data Preporessing and then fed into the front-end Tracking module. In the front-end, the system employs Gaussian ellipsoids rendered from the previous time instant in conjunction with the current point clouds for frame-to-map matching. Keyframes are evaluated through the co-visibility in visual regions and are appended to the Gaussian map. The Gaussian map incorporating keyframe parameters is then processed in the back-end for pose optimization and map updates. By minimizing the loss function, the Gaussian map updates the parameters of Gaussians continuously together with splitting and pruning operations of Gaussians.

% In simpler terms, throughout the SLAM process, the system incorporates information from the current frame into the Gaussian map based on localization results. The back-end loss function optimizes the trajectory and environmental representation through backpropagation, based on continuous changes in keyframes, thereby achieving optimal parameter adjustments for localization and map construction.

In the proposed system, data inputs consist of imagery from a camera and point clouds from a LiDAR sensor. These inputs are integrated using calibrated extrinsic to transform the time-aligned LiDAR point clouds into depth images.

The transformation is governed by the equation:

\begin{equation}
    Depth = K \cdot (R^{C}_{L} \cdot P + t^{C}_{L})
\end{equation}
where \( P \) represents the points in the LiDAR point clouds, \( R^{C}_{L} \) and \( t^{C}_{L} \) denote the rotation matrix and translation vector from the LiDAR to the camera coordinate system, respectively, and \( K \) is the intrinsic matrix of the camera. This transformation aligns the LiDAR data with the visual data, facilitating a unified depth perception from the RGB viewpoint.

\subsection{Front-End Tracking}
The primary challenge in front-end tracking arises from how to align variably dense Gaussians derived from color supervision with LiDAR point clouds that possess a fixed resolution. Inspired by \cite{ha2024rgbdgsicpslam}, \cite{guedon2024sugar} and \cite{behley2018SUMA}, we adopt covariance, a shared attribute of point clouds and Gaussians, as a bridge for integration. After determining the normal orientation for each LiDAR point, we align it with the shortest axis of the Gaussians. To further facilitate stable tracking, we introduce a weighting function for Gaussians that distinguishes Gaussians generated solely by color supervision and those also by LiDAR depth in the error calculation of point clouds and Gaussian match. 

% Since the implicit of the gradient descent process, coupled with the sparsity and discreteness inherent in LiDAR point clouds in expansive outdoor settings, make direct covariance integration suboptimal. However, the orientation of normals has been consistently identified as a more robust factor in outdoor environments within the scope of LiDAR-based SLAM research. 
\textbf{Point clouds and Gaussians match:} Initially, we maintain a sliding window that filters and selects Gaussians from the most recent 10-time frames while masking out the remaining Gaussians. This selection process keeps Gaussians relevant for matching within our focused sub-maps. We then utilize an incremental error minimization function to ensure precise correspondences between planes and points as  below:
\begin{equation}\label{p&p icp}
    E(x_p, x_g) = \sum_{x_p \in P_{L}} w_{x_g}  (n_{x_p} \cdot ({T^{C_{t-1}}_{W}}^{(k)} x_p - x_g)^2) 
+\text{R}
\end{equation}
where \(x_p\) represents a point in the LiDAR point clouds \(P_{L}\), \(T^{C_{t-1}}_{W}{}^{(k)}\) corresponds to the current pose estimation after \(k^{th}\) iterations based on the pose from the previous moment to the world coordinate system and \(x_g\) denotes the center of the Gaussian closest to \( T^{C_{t-1}}_{W}{}^{(k)}x_p \), and \(n_{x_p}\) is the normal vector of \(x_p\). \(w_{x_g}\) is a weight representing the confidence of the point \(x_g\), which is detailed in the next paragraph.

% Gaussians may change after being adjusted by the color loss function. To ensure reliable matching results, it is essential to use reliable points for matching. Therefore, we introduce a weighting mechanism to ensure point reliability. Ellipsoids derived from color supervision lack LiDAR depth constraints and may fill the image with large ellipsoids without geometric constraints. Regions frequently receiving depth information tend to have higher density. Consequently, the reliability of a point is proportional to its density and the consistency of its normal trend.

The regularization term $R$ is introduced to further enhance the stability and precision of error funcition, which considers the directional error between normals:
\begin{equation}
    \text{R} = \left\| 1 - n_{x_p} \cdot n_{x_g} \right\|
\end{equation}

% The regularization term would be multiplied by a hyperparameter for adjustment, aims to minimize the distance between each point pair, while simultaneously reinforcing the alignment of normals. 
The regularization term would be multiplied by a hyperparameter for adjustment, aims to reinforce the alignment of normals. 

\begin{figure}[t]
    \centering
    \includegraphics[width=0.9\linewidth]{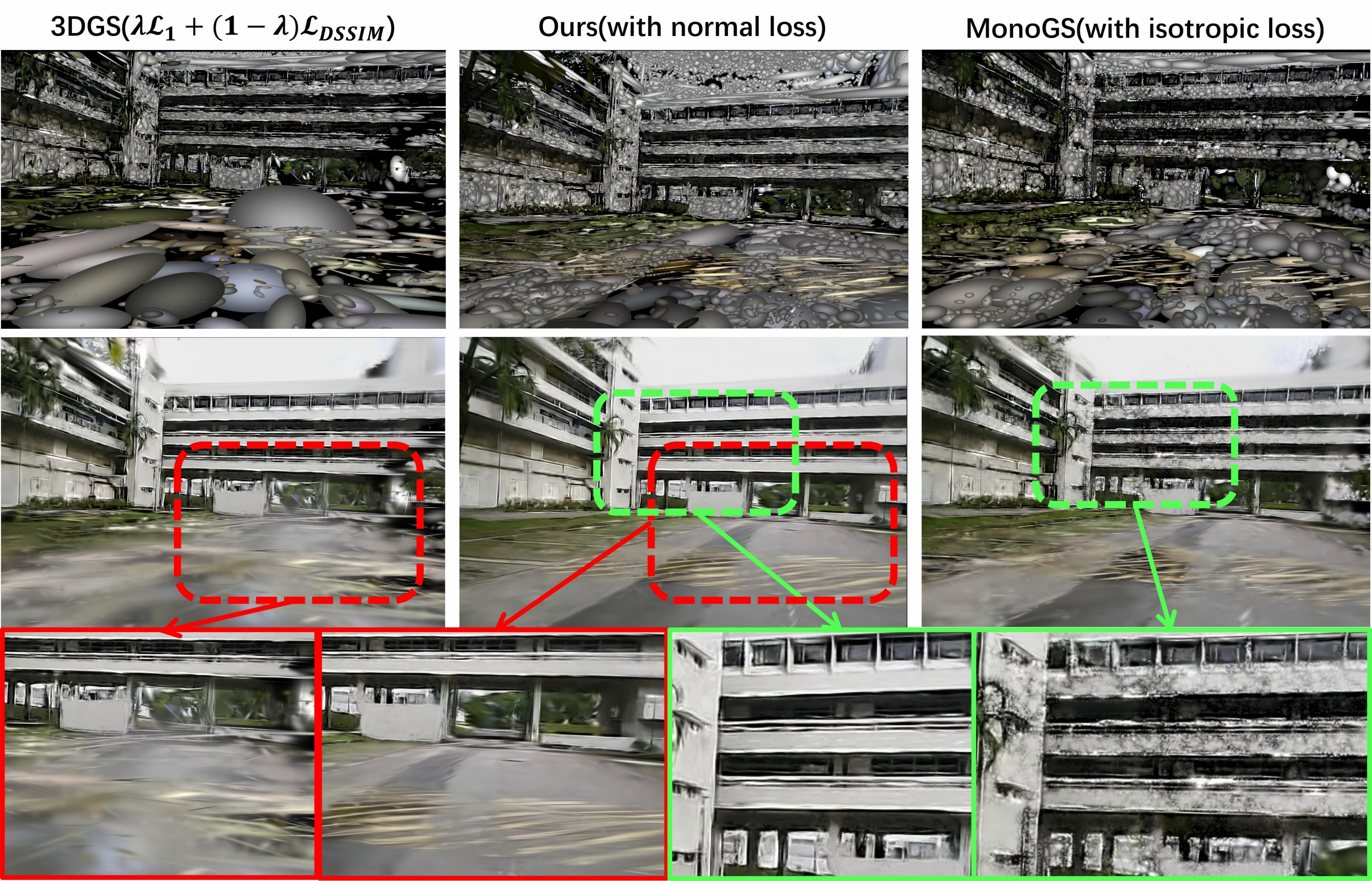}
    \caption{\textbf{Effect of Normal Restriction:} \textbf{Top:} Ellipsoid visualization. \textbf{Middle:} Render images. \textbf{Bottom:} Magnified details of the render. The left comparison (in red) illustrates uncontrolled Gaussian growth leading to significant artifacts. The right comparison (in green) shows gaps in the rendered images caused by isotropic Gaussians. Our method effectively prevents these issues.}
    \label{loss}
\vspace{-3mm}
\end{figure}

\textbf{Weighting function construction:} The reliable weight of the point clouds is closely linked to its density, as depicted in Fig. \ref{dense}. To construct a weight function that combines normal consistency and density factors, we first identify all the nearest Gaussian distribution centers inside the local spherical region $N_r(x)$,  where  \(x\) is the center of the sphere and $r$ is the radius. These Gaussian centers are determined by minimizing the distance from point \(x\) to each Gaussian center \(\mu_{g_i}\). The density function \(\rho(x)\) is given below:

\begin{equation}\label{density}
\rho(x) = \sum_{g_i \in N_r(x)} \alpha_i \exp \left( -\frac{1}{2} (x - \mu_{g_i})^T {\Sigma'_{g_i}}^{-1} (x - \mu_{g_i}) \right)
\end{equation}
\begin{equation}
\Sigma'_{g_i} = D_{g_i} \, \text{diag}(1, \sigma_{\text{perp1}}/\sigma_{\text{along}}, \sigma_{\text{perp2}}/\sigma_{\text{along}}) \, D^T_{g_i}
\end{equation}
where \(\Sigma'_{g_i}\) is the reconstructed covariance matrix, achieved by selecting the smallest variance \(\sigma_{\text{along}}\) along the normal direction and a larger variance \(\sigma_{\text{perp}}\) in the perpendicular directions. \(D_{g_i}\) is an orthogonal basis formed by the normal and its perpendiculars. Since the length of the Gaussian normal is difficult to restrict during the optimization, we introduced the normal length normalization for both point clouds and Gaussians, so that the robustness of the tracking algorithm can be ensured by the stable orientation of the normal. 

In addition, since \eqref{density} involves calculations of the density of Gaussians per point during the matching process, to speed up the calculation process, we simplify the density function calculation in \eqref{density}  while tracking:
\begin{equation}
    (x - \mu_g)^T {\Sigma_g'}^{-1} (x - \mu_g) \approx \frac{\sigma_{\text{perp1}}\sigma_{\text{perp2}}}{\sigma_{\text{along}}^2} \langle x - \mu_g, n_{x_g} \rangle^2
\end{equation}

For each point \(x\), we then calculate the consistency \(C(x)\) between the normal \(n_{x_g}\) of the current Gaussian distribution and local average normal \(\bar{n}\) as
   $ C(x) = n_{x_g} \cdot \bar{n}$. The final weight function \(W(x)\) is defined as the product of the normal consistency and the density function $
    W(x) = C(x)\rho(x)$.

% This approach ensures that the weight is high only in regions where the normals are highly consistent and the density is high. 

\textbf{Co-visibility assessment:} We measure co-visibility by assessing the overlap of Gaussian functions between the current and last keyframe. Should co-visibility fall below a specified threshold, the frame is designated as a keyframe. To guarantee that fast tracking under a forward-looking perspective consistently matches the current point clouds with sufficient Gaussian ellipsoids, point clouds from the keyframes are added to the Gaussian map directly, using pixel colors and scaling point clouds covariance along the ray propagation direction for initialization. Subsequently, the back-end optimizes the augmented Gaussian map by comparing it against the perspectives of previous keyframes, leading to incremental updates of the Gaussian map.

\begin{figure}[t]
    \centering
    \includegraphics[width=0.9\linewidth]{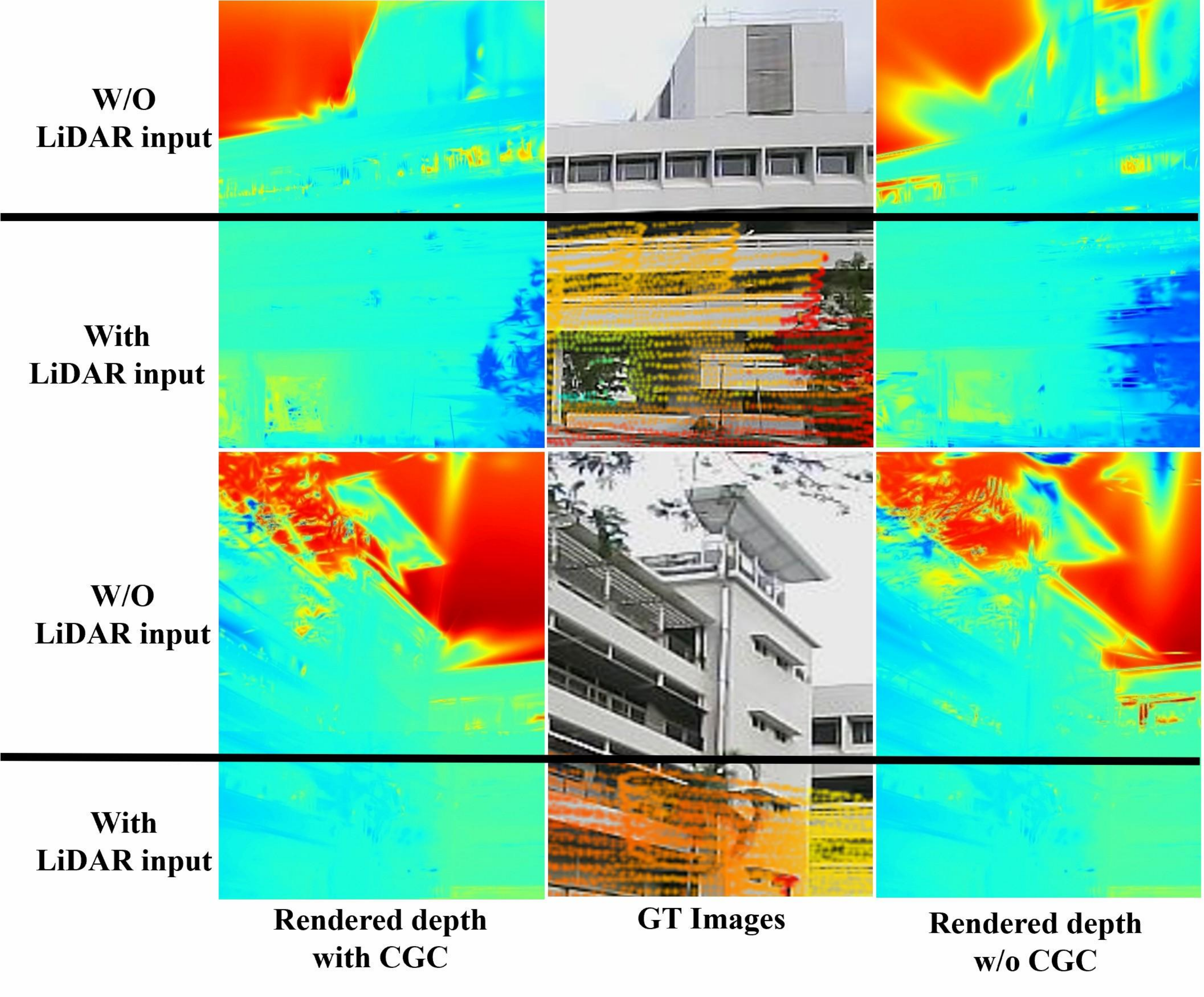}
    \caption{\textbf{Effect of Splitting via conditional Gaussian constraints (CGC). }Our approach enhances the representation of Gaussians  for objects in the images that lack LiDAR depth input via the introduced  CGC.}
    \label{conditial gaussian}
\end{figure}

\begin{table*}[t]
  \caption{Quantitative Analysis for Tracking Accuracy}
  \label{traj}
  \centering
  \resizebox{\textwidth}{!}{
  \begin{tabular}{@{}lccc|ccc|ccc|ccc|ccc|ccc@{}} % 正确的列定义：1列Methods + 6个数据集*每个3列
    \toprule
     \multicolumn{1}{c}{\multirow{3}{*}{Methods}} & \multicolumn{3}{c}{cp} & \multicolumn{3}{c}{garden1} & \multicolumn{3}{c}{garden2} & \multicolumn{3}{c}{nyl1} & \multicolumn{3}{c}{nyl2} & \multicolumn{3}{c}{loop2} \\
    \cmidrule(lr){2-4} \cmidrule(lr){5-7} \cmidrule(lr){8-10} \cmidrule(lr){11-13} \cmidrule(lr){14-16} \cmidrule(lr){17-19}
    & $t_{rel}\downarrow$ & $r_{rel}\downarrow$ & $t_{abs}\downarrow$ & $t_{rel}\downarrow$ & $r_{rel}\downarrow$ & $t_{abs}\downarrow$ & $t_{rel}\downarrow$ & $r_{rel}\downarrow$ & $t_{abs}\downarrow$ & $t_{rel}\downarrow$ & $r_{rel}\downarrow$ & $t_{abs}\downarrow$ & $t_{rel}\downarrow$ & $r_{rel}\downarrow$ & $t_{abs}\downarrow$ & $t_{rel}\downarrow$ & $r_{rel}\downarrow$ & $t_{abs}\downarrow$ \\
    \midrule
    NeRF-LOAM & 2.943 & 9.644 & 5.39 & 1.182 &\textbf{0.559} & 0.540 & 1.213 &\textbf{0.707} & 1.076 & 1.371 & \textbf{1.140} & 3.504 & 1.343 & 1.730 & 17.460 & 1.442 & \textbf{2.205} & 1.785\\
    HDL-graph-SLAM & 1.264 & 1.553 & 1.079 &1.874 &1.603 &1.478  &\textbf{1.186}  &0.880 & 3.154 &1.737 &1.271 & \underline{2.266} &1.514  &1.835  & 17.638 &1.436  &2.802  &\textbf{0.593}\\
    ORB-SLAM3 & 1.356 & 1.992    & 2.865    & \textbf{1.173} & \underline{0.626}    & \underline{0.529}    & \underline{1.212} & \underline{0.772}    & 1.001   & \underline{1.342} & \underline{1.172}   & 19.528 &\underline{1.333} & 1.736   & 23.283  & \underline{1.403} & 2.256   & 0.952   \\
    SplaTAM &  - &  -    & 2.336    & - &  -    & 0.979    & - & -    & 1.221   & - & -    & 12.332 &- & -    &\underline{17.442}   & - &  -    & 2.692   \\
    MonoGS & 4.171 & 3.472 & 3.440 & \underline{1.179} &0.754 & 0.664 & 1.163 & 0.765 & \underline{0.708} &1.382 & 1.175 & 9.595 & 1.371 &\underline{1.701} & 28.553 & 7.375 &5.708 & 15.357 \\
    Gaussian-SLAM & \underline{1.249} & \underline{3.047} & \underline{1.040} & - & - & - & - & - & - & - & - & - & - & - & - & \underline{1.399} & 2.384 & 1.136 \\
    GS-ICP-SLAM & 5.471 & 4.041 & 6.33 & 1.249 & 0.764 & 2.082 & 1.824 & 1.316 & 5.507 & 1.662 & 1.771 & 23.331 & 2.101 & 1.070 & 23.915 & 3.236 & 2.644 & 13.819 \\
    Ours & \textbf{0.234} & \textbf{1.216}   & \textbf{0.464}   & 1.183 &0.716   & \textbf{0.366}    & 1.236 &0.962    & \textbf{0.679}   & \textbf{1.240} &1.307   & \textbf{0.580}    & \textbf{1.106} & \textbf{1.369}    & \textbf{0.771}   & \textbf{1.393} &\underline{2.239}   & \underline{0.843}   \\
    \bottomrule
  \end{tabular}
}
\end{table*}

\begin{table*}[t]
\centering
\caption{Quantitative Analysis for Rendering [SSIM$\uparrow$ PSNR$\uparrow$ LPIPS$\downarrow$]}
\label{render}
\resizebox{\textwidth}{!}{% 缩放表格以适应文本宽度
\begin{tabular}{@{}lcccccccccccccccccc@{}}
\toprule
& \multicolumn{3}{c}{\textbf{cp}} & \multicolumn{3}{c}{\textbf{garden1}} & \multicolumn{3}{c}{\textbf{garden2}} & \multicolumn{3}{c}{\textbf{nyl1}} & \multicolumn{3}{c}{\textbf{nyl2}} & \multicolumn{3}{c}{\textbf{loop2}}  \\
\cmidrule(r){2-4} \cmidrule(lr){5-7} \cmidrule(lr){8-10} \cmidrule(l){11-13} \cmidrule(l){14-16}  \cmidrule(l){17-19}
\textbf{Methods} & \textbf{SSIM} & \textbf{PSNR} & \textbf{LPIPS} & \textbf{SSIM} & \textbf{PSNR} & \textbf{LPIPS} & \textbf{SSIM} & \textbf{PSNR} & \textbf{LPIPS} & \textbf{SSIM} & \textbf{PSNR} & \textbf{LPIPS}  & \textbf{SSIM} & \textbf{PSNR} & \textbf{LPIPS}  & \textbf{SSIM} & \textbf{PSNR} & \textbf{LPIPS} \\
\midrule
3DGS                      & 0.718 & 21.95& 0.617 & 0.598 & 20.495 & 0.557 & 0.662 & 20.402 & 0.536 &  0.726 & 20.224 & 0.575 & 0.571 & 18.040 & 0.657 & 0.594 &  17.794 & 0.620\\

GS-ICP-SLAM              & 0.552 & 17.336 & 0.772 & 0.511 & 15.782 & 0.661 & 0.417 & 13.946 & 0.606 &  - & - & - & 0.374 & 11.533 & 0.882&  0.492 & 12.335 & 0.771\\

NeRF++                    & 0.566 & 19.226 & 0.698 & 0.529 & 15.772& 0.685 & 0.475 & 13.736 & 0.675& 0.574 & 13.662 &  0.726 & 0.320 & 10.130 & 0.879 & - & -  & -\\

SplaTAM (Odom)            & 0.629 & 18.772& 0.585 & 0.515 & 18.788 & 0.545 & 0.540 & 18.402 & 0.569 & 0.688 & 19.770 & 0.574 & 0.480 & 17.212 & 0.644 & 0.475 & 13.736 & 0.675\\

SplaTAM (GT)              & 0.535 & 17.501& 0.691 & 0.498 & 17.936 & 0.607 & 0.497 & 18.402 & 0.533 & 0.597 & 18.960 & 0.520 & 0.432 & 16.933 & 0.721 &0.438 & 11.225 & 0.753\\

MonoGS (Odom)             & 0.655 & 20.962 & 0.444 & 0.607 & 23.124 & 0.474 & 0.649 & 20.627 & 0.535 & \underline{0.813} &  \textbf{22.800} & 0.373 & 0.560 & 17.812 & 0.665 & 0.578 & 17.993 &0.503\\
MonoGS  (GT)              & 0.733 & 22.118 & 0.364 & 0.717 & \textbf{23.723} & 0.522 & 0.598 & 20.112 & 0.552 & 0.748 &  21.969 & 0.415 & 0.511 & 17.006 & 0.738 & 0.505 & 17.567 &0.616\\

Gaussian-SLAM (Odom)      & 0.665 & 20.924& 0.595 & - & - & - & - & - & - & - & - & - & -     & -    &- & 0.534 & 16.509 & 0.656\\
Gaussian-SLAM (GT)        & 0.647 & 21.026& 0.688 & - & - & - & - & - & - & - & - & - & -     & -    &- & - & - & -\\

Ours (Odom)               & \textbf{0.775}& \underline{22.274} & \underline{0.336} & \textbf{0.773} & \underline{23.569} & \textbf{0.386} & \textbf{0.772} & \textbf{22.141} & \textbf{0.336} & \textbf{0.833} & \underline{22.559} & \textbf{0.274} & \underline{0.725} & \textbf{20.835} & \underline{0.559} & \textbf{0.686} & \underline{18.037} & \underline{0.556}\\
Ours (GT)                 & \underline{0.763} & \textbf{22.368} & \textbf{0.319} &  \underline{0.721} & 23.353 & \underline{0.404} & \underline{0.758} & \underline{21.991} & \underline{0.348} & 0.799 & 22.136 & \underline{0.308} & \textbf{0.744} & \underline{20.630} & \textbf{0.548} & \underline{0.631} & \textbf{18.472} & \textbf{0.417}\\
\bottomrule
\end{tabular}
}
\end{table*}

\subsection{Back-End Optimization}
The back-end optimization process retrieves a sequence of keyframe identifiers along with their corresponding parameters and conducts two rounds of optimization. The first time is confined to the poses of keyframes within the sliding window, while the second time is aimed at updating the Gaussian map. The depth and color rendering process from the 3D Gaussian map \( G^s \) is derived as:
\begin{equation}
D_{\text{render}} = \sum_{g_i \in G^s} d_i \alpha_i \prod_{j=1}^{i-1} (1 - \alpha_j)
\end{equation}
\begin{equation}
C_{\text{render}} = \sum_{g_i \in G^s} c_i \alpha_i \prod_{j=1}^{i-1} (1 - \alpha_j)
\end{equation}
where \(d_i\) and \(c_i\) represent the distance and color to the Gaussian \(g_i\) along the camera ray. 

The loss function used for optimizing the parameters of Gussians is designed as:
\begin{equation}\label{loss-func}
\mathcal{L} = (1 - \lambda_1) E_{\text{pho}} + \lambda_1 E_{\text{geo}} + \lambda_2 E_{\text{normal}}
\end{equation}
where the first two terms are commonly used in existing research works: the photometric error \(E_{\text{pho}}\) represents the discrepancy between the visual ground truth and the rendered image, while the geometric error \(E_{\text{geo}}\) measures the disparity between the LiDAR depth input and the rendered depth image. The third term $E_{\text{normal}} = \|\overline{\sigma_{\text{along}}}\|$ quantifies the norm of the average variance along the normal direction.

% This function is integrated into the sliding window of keyframes in the backend. Considering the unpredictability of Gaussian scale changes during supervised optimization, unlike indoor environments relying on dense, continuous depth from RGB-D cameras for constructing heavily stacked Gaussians, outdoor scenes typically feature larger and fewer Gaussians. We aim for the Gaussians to closely align with the object surfaces, possessing shorter and more stable normals. To this end, we introduce a constraint that minimizes the global normals.

% \input{table/traj}
% \input{table/render}

% This constraint enhances the stability of the Gaussian map, significantly benefiting the stability of the normal orientation within the front-end tracking component. Additionally, by supervising normals, it ensures that Gaussians which initialized based on the covariance of the point clouds closely conform to the object surfaces. This adherence strengthens the spatial structure of the Gaussian map, as shown in Fig. \ref{loss}, ultimately improving the rendering quality.
Unlike indoor 3D Gaussian splatting where stacked Gaussians are constructed from dense RGB-D images, outdoor scenes typically feature larger and sparser Gaussians. To align Gaussians with object surfaces closely, we introduce the normal loss $E_{\text{normal}}$ which is optimized for shorter and more stable normals. As demonstrated in Fig. \ref{loss}, with the introduced normal loss, the 3D Gaussian map are structured with more stable normals, improving rendering quality greatly.

% \begin{equation}
% E_{\text{normal}} = \sum_{i \in G^s}\|\overline{\sigma_{\text{along}}}\|
% \end{equation}

\subsection{Map Update}
% Through extensive experiments and observations, we have found that points closer to the object surface are geometrically more stable and exhibit minimal color variation. Thus, we have adopted an innovative approach by substituting the traditional Gaussian covariance with the modified covariance derived from LiDAR point clouds. This substitution introduces new geometric constraints to the Gaussian Gaussians, effectively supervising the splitting of Gaussians during the gradient descent process. Unlike traditional approaches that rely solely on LiDAR point clouds for map construction and visual inputs for pose estimation, each module of our system interacts with the 3D Gaussian map, enhancing both localization accuracy and mapping fidelity.
To manage the representation of the sky in unbounded scenes, we adopt a method similar to \cite{kerbl2024hierarchical}, initializing a skybox outside the scene with 100,000 Gaussian primitives. This skybox is dynamically updated as the Gaussian map is incrementally built. During map initialization and update, two types of Gaussians are predominant: Gaussians derived from color supervision \(X\) and those provided by LiDAR measurements \(Y\). 

We introduce a Conditional Gaussian Constraint (CGC) to adjust the positions of color-supervised Gaussians through the loss function \eqref{loss-func}. For each point \(x\) acquired through color supervision, the nearest Gaussian \(y\) is selected from the LiDAR-measured Gaussians set. Further, it is postulated that  given \(y\), \(x\) follows a Gaussian distribution:
\begin{equation}\label{condition-Gaussian}
p(X | Y = y )\sim \mathcal{N}(\mu_x(y), \Sigma_{y})
\end{equation}
where $\mu_x(y)$ is the location of new Gaussian ellipsoid split from Gaussian $y$ through normal distribution sampling, and $\Sigma_{y}$ is the covariance of Gaussian $y$.

The conditional Gaussian equation \eqref{condition-Gaussian} adjusts the mean \(\mu_x\) and covariance \(\Sigma_x\) of \(x\), aligning them more closely with the nearest reliable Gaussian $y$. The new Gaussian split from the reliable Gaussian  is regarded as reliable Gaussian after undergoing a round of back-end optimization, the process of which continue until all Gaussians become reliable Gaussians. As shown in Fig. \ref{conditial gaussian}, the newly-split points will strictly adhere to the distribution patterns of existing reliable points, especially in areas with complex shapes or distinctive surface features.
% \begin{equation}
% \mu_{x|y} = \mu_x + \Sigma_{xy} \Sigma_{y}^{-1} (y - \mu_y)
% \end{equation}
% \begin{equation}
% \Sigma_{x|y} = \Sigma_{x} - \Sigma_{xy} \Sigma_{y}^{-1} \Sigma_{yx}
% \end{equation}
% Here, \(\Sigma_{x}\) represents the original covariance of \(x\), \(\Sigma_{y}\) the covariance of \(y\), and \(\Sigma_{xy}\) the covariance between \(x\) and \(y\).

% This method implies that the distribution of new Gaussians will directly simulate the local characteristics of the nearest reliable Gaussian. By adjusting the color-supervised Gaussians to more accurately reflect their physical properties, the consistency of the Gaussians is enhanced. The newly splited points will strictly adhere to the distribution patterns of known points, especially in areas with complex shapes or distinctive surface features.

\begin{figure}[t]
\centering
\subfloat[cp\label{01}]{
    \includegraphics[width=0.45\linewidth]{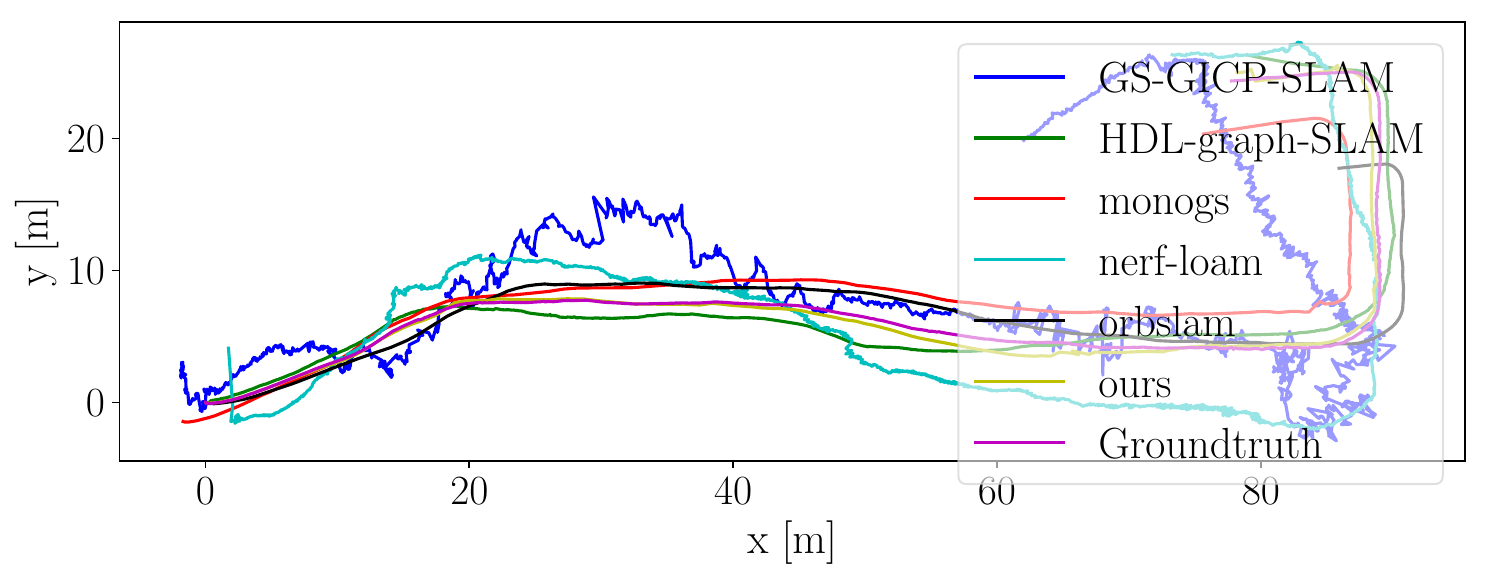}}
\subfloat[nyl1\label{02}]{
    \includegraphics[width=0.45\linewidth]{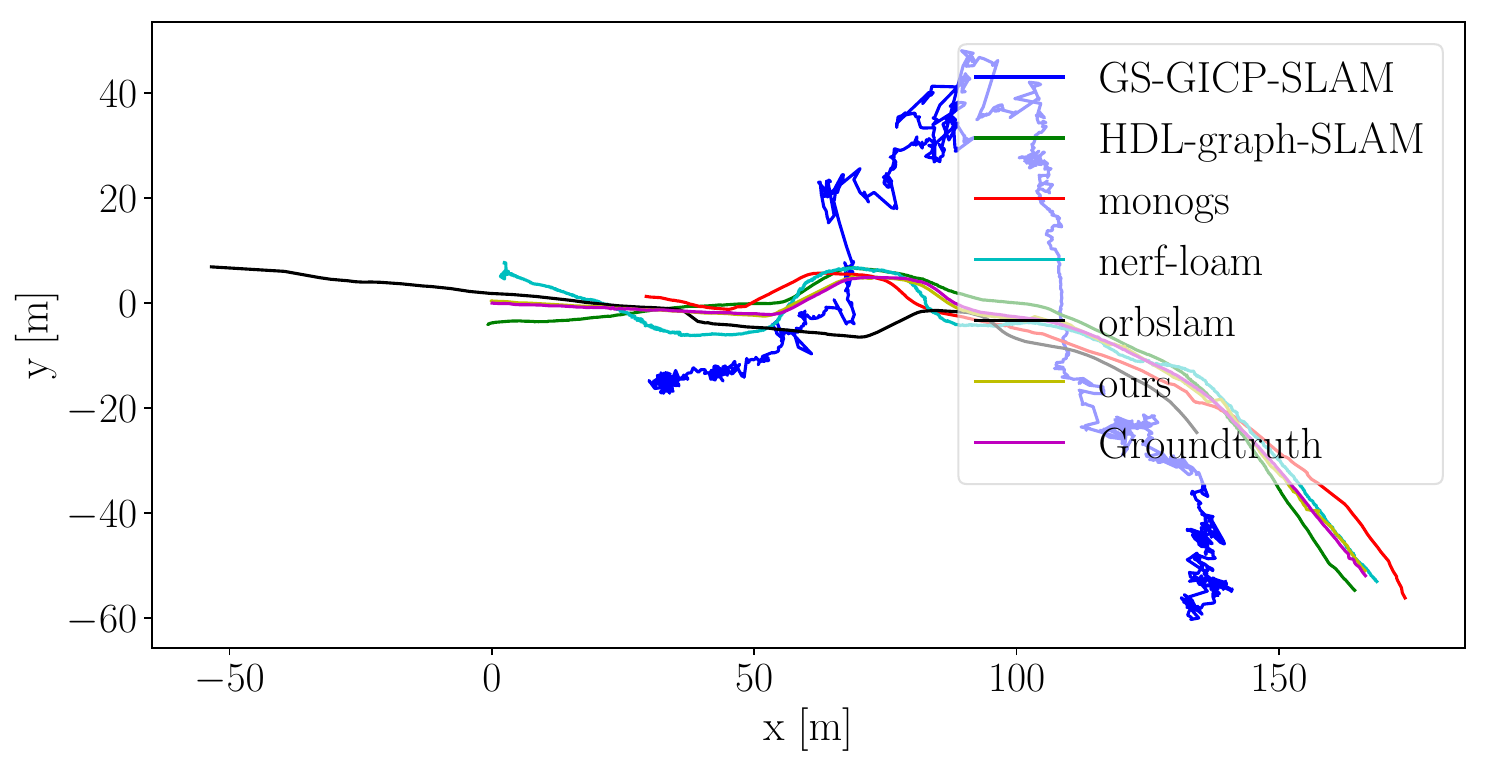}} \\
\subfloat[nyl2\label{03}]{
    \includegraphics[width=0.45\linewidth]{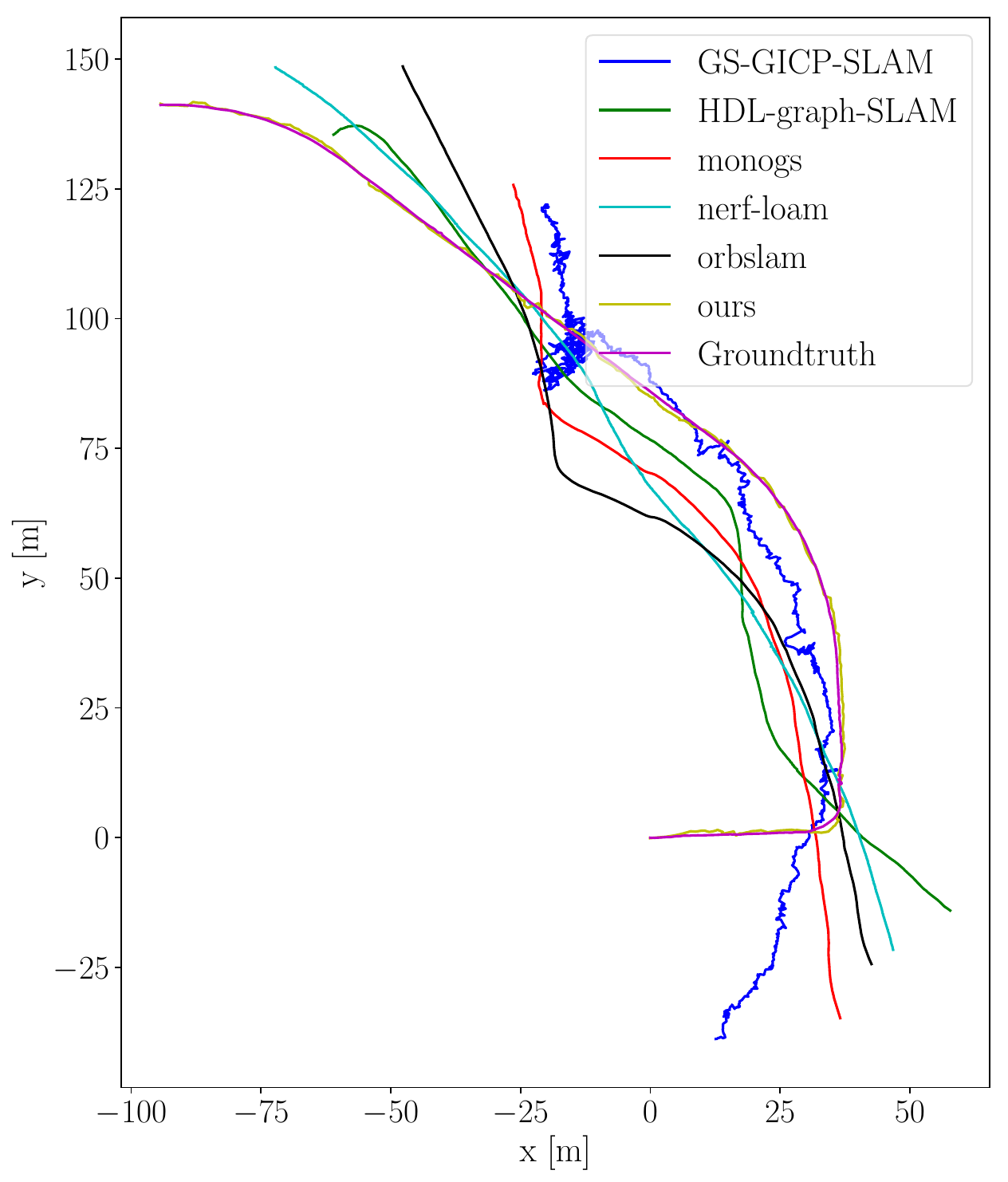}}
\subfloat[loop2\label{04}]{
    \includegraphics[width=0.45\linewidth]{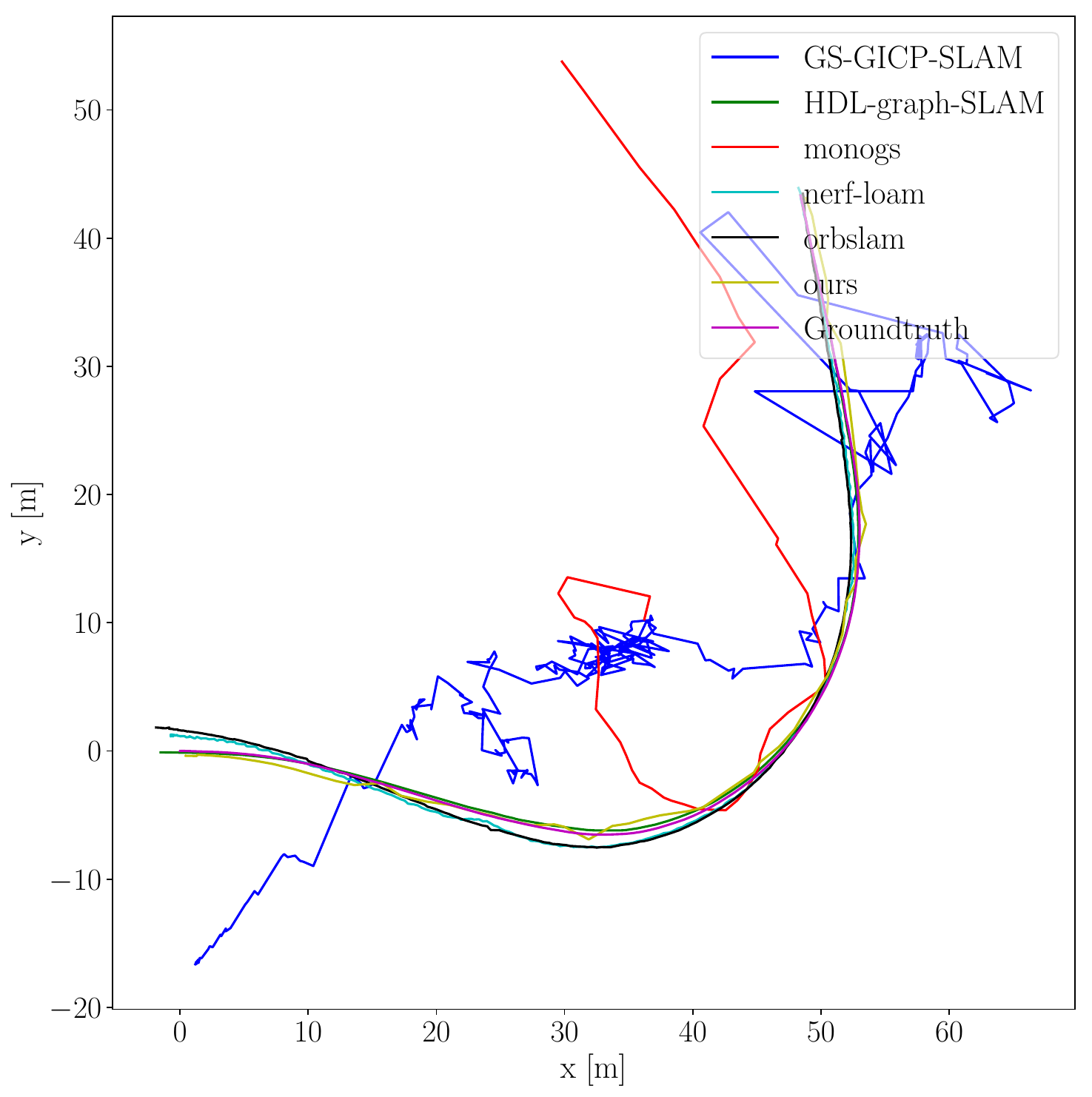}}
\caption{\textbf{Comparison of trajectories using different SLAM algorithms on four sequences of NTU4DRadLM dataset.}}
\label{top view}
\vspace{-3mm}
\end{figure}

\begin{figure*}[htbp]
    \centering
    \includegraphics[width=\linewidth]{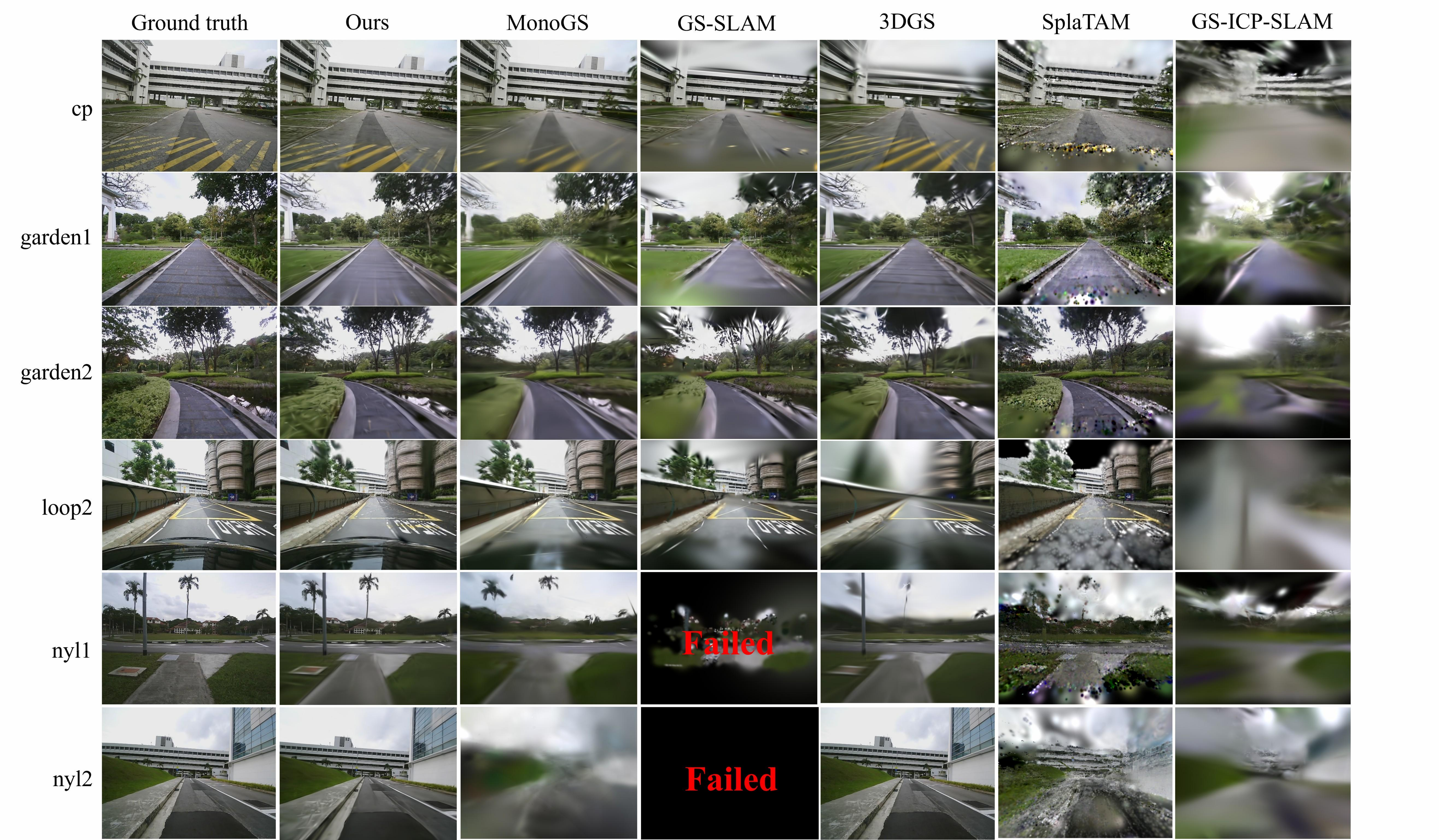}
    \caption{\textbf{Comparison of Rendering Results.}}
    \label{Comparison of Render}
\vspace{-3mm}
\end{figure*}

\section{Experiment}
In the experiments,  we evaluate  LiV-GS and compare it against other SOTA algorithms from three aspects: localization accuracy, rendering quality, and the reliability of spatial distribution of Gaussian maps.

To evaluate trajectory error, we used the open-source tool rpg trajectory evaluation\cite{Zhang18irosrpg} to compute both Absolute Trajectory Error (ATE) and Relative Error (RE), measuring the ATE root-mean-square error (RMSE) drift (m), average translational RMSE drift (\%) and average rotational RMSE drift (°/100 m). 

% As to the spatial geometric distribution of the Gaussian map, we conducted a cross-modal mmWave radar localization on the produced Gaussian maps, as the more reliable in spatial geometric of the Gaussian map, the easier the cross-modal localization is. 

For rendering evaluation, the optimized viewpoints from each algorithm were extracted and compared against the actual images using metrics of SSIM, PSNR[dB], and LPIPS.
% \begin{figure*}[htbp]
%     \centering
%     \includegraphics[width=\linewidth]{figures/pdf/render show.pdf}
%     \caption{\textbf{Comparison of Rendering Results.}}
%     \label{Comparison of Render}
% \end{figure*}

\subsection{Implementation Details}
LiDAR and image data were synchronized using timestamps, and the trajectories obtained by the R$^{3}$Live\cite{lin2022r3live} algorithm which integrates vision, LiDAR, and IMU data, is used as the ground truth. As the trajectory accuracy, LiV-GS is compared with the established point clouds-based geometric SLAM algorithm HDL graph SLAM \cite{koide2019hdl}, image feature-based visual SLAM algorithm ORB-SLAM3 \cite{campos2021orbslam3}, and implicit neural field-based NeRF-LOAM \cite{deng2023nerfloam}, and the open-source SLAM algorithms based on 3D Gaussian Splatting such as MonoGS \cite{matsuki2024gaussianslams}, Gaussian-SLAM \cite{yugay2023gaussianslam}, GS-ICP-SLAM\cite{ha2024rgbdgsicpslam} and SplaTAM \cite{keetha2024splatam}. Image rendering quality comparisons are performed under gtpose and its odometry using algorithms including 3DGS, NeRF++\cite{zhang2020nerf++}, MonoGS, Gaussian-SLAM, GS-ICP-SLAM and SplatTAM. All of the algorithms were run on a desktop with an NVIDIA RTX 4090 GPU.

\subsection{Datasets}
To effectively evaluate our LiV-GS, we utilized the open-source large-scale dataset NTU4DRadLM, which includes the data collected by three different types of sensors: Livox-Horizon LiDAR at 10Hz, a monocular camera with a resolution $640\times480$, and a 4D millimeter wave radar Eagle Ocuill G7. Our method does not use the IMU data. Due to the challenge of maintaining photometric consistency in long-distance outdoor scenes, we segmented the low-speed, kilometers-long scenarios as several shorter sequences. For the cp sequence, we used the first 2400 LiDAR-camera aligned images, covering approximately 230 meters. For the garden and nyl sequences, we selected 2100 and 2400 images respectively from both the beginning and end of each sequence, with each segment covering at least 220 meters. Additionally, for the Loop2  sequence recorded on a human-driving vehicle platform, we selected 300 frames covering about 250 meters. In total, we tested six sequences to comprehensively validate our LiV-GS method.

% \input{table/traj}

% To effectively evaluate our LiV-GS method, we utilized the open-source large-scale datasets \textbf{NTU4DRadLM} and \textbf{R3Live}. The NTU4DRadLM dataset includes data from a 10\,Hz \textit{Livox-Horizon} LiDAR and a $640 \times 480$ camera, while the R3Live dataset uses a 10\,Hz \textit{Livox-Avia} LiDAR and images with a resolution of $1280 \times 1024$. It is important to note that our method does not incorporate IMU data input.

% Due to the challenges of maintaining photometric consistency in long-distance outdoor scenes, we divided the low-speed, multi-kilometer scenarios in NTU4DRadLM into several shorter sequences. For the \textbf{cp} sequence, we used the first 2,400 LiDAR-camera aligned images, covering approximately 230 meters. For the \textbf{garden} and \textbf{nyl} sequences, we selected 2,100 and 2,400 images from the beginning and end of each sequence, respectively, each covering at least 220 meters. Additionally, for the \textbf{Loop2} sequence recorded on a manned vehicle platform, we selected 300 frames covering about 250 meters.

% In the R3Live dataset, we extracted a looped sequence from \textbf{hku\_park\_00}. We tested a total of seven sequences to comprehensively validate our LiV-GS method.

\subsection{Evaluation of Tracking Accuracy}
Fig. \ref{top view} and Table \ref{traj} display qualitative and quantitative evaluations of localization accuracy, respectively. Our LiV-GS demonstrates the lowest ATE in low-speed sequences. However, in high-speed sequence loop2, the accuracy of LiV-GS is slightly lower than NeRF-LOAM due to larger displacement between the consecutive frames and sparse Gaussian distribution led by insufficient Gaussian map optimizations. MonoGS, SplaTAM, GS-ICP-SLAM, and Gaussian-SLAM are all tailored for indoor environments with well-textured images and dense depth information, and they suffer performance degradation or even fail in some outdoor sequences due to sparser depth information obtained in outdoor scenes. In contrast, our LiV-GS tracks consistently and stably in large-scale outdoor environments.

% Our method leverages the relationship between LiDAR point clouds and Gaussians for robust pose estimation, minimizing scale drift. Additionally, the covariance of previous keyframe point clouds is used as a prior to generate new Gaussians, increasing the density of reliable ellipsoids and enhancing the reliability of the matching process. Nerf-LOAM is developed for high-speed mobility with 360-degree LiDAR but is limited by frontward data due to its projection onto a camera view.

In addition, the bottom two rows in Table \ref{render} show a minimal rending difference between the outcomes from LiV-GS odometry and ground truth, which further justifies the high precision of LiV-GS odometry, as higher localization accuracy leads to a smaller loss in rending quality.

% \input{table/render}
% \begin{figure}[htbp]
% \centering
% \subfloat[Comparison of image quality and running speed in cp\label{02}]{
%     \includegraphics[width=\linewidth]{figures/FPS.png}}\\
%     \label{Image quality}
% \subfloat[Comparison of localization accuracy and running speed in nyl\label{04}]{
%     \includegraphics[width=\linewidth]{figures/FPS-accuracy.png}} 
%     \label{absolute error}
% \caption{\textbf{Comparison of evaluation metrics relative to the system runtime.} Our approach delivers state-of-the-art performance in terms of rendering quality and system-wide localization accuracy.}
% \label{Comparison of FPS}
% \end{figure}

% \begin{figure*}[htbp]
%     \centering
%     \includegraphics[width=\linewidth]{figures/pdf/render show.pdf}
%     \caption{\textbf{Comparison of Rendering Results.}}
%     \label{Comparison of Render}
% \end{figure*}

\begin{figure}[t]
\centering
\subfloat[Comparison of image quality and running speed in cp\label{Image quality}]{
    \includegraphics[width=0.9\linewidth]{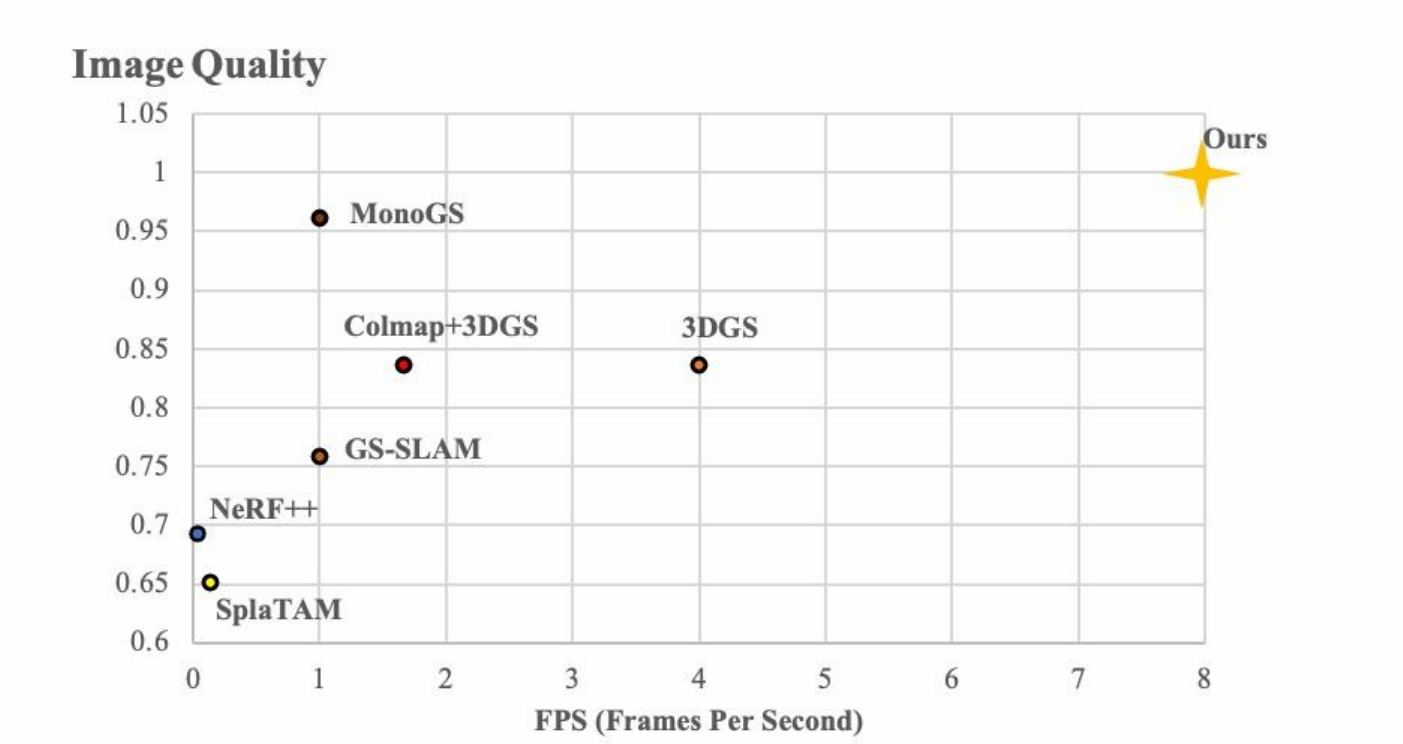}}
\\
\subfloat[Comparison of localization accuracy and running speed in nyl\label{absolute error}]{
    \includegraphics[width=0.8\linewidth]{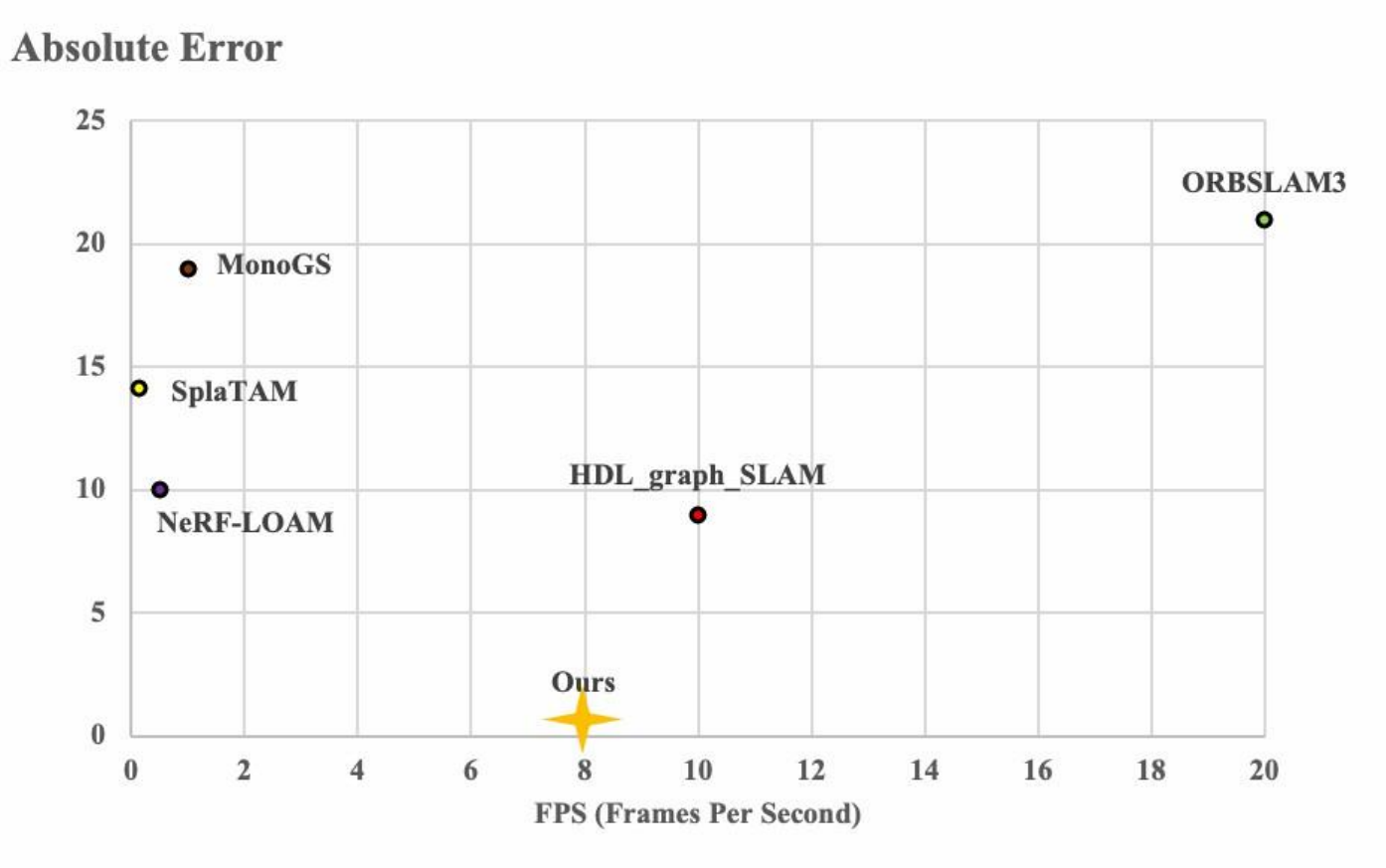}} 
\caption{\textbf{Comparison of performance metrics and system runtime.}  FPS is calculated as the ratio of the total number of processed frames to the total time. A value of 7.98 FPS means that on average 7.98 depth maps and RGB images can be processed in one second.}
\label{Comparison of FPS}
\vspace{-5mm}
\end{figure}

\subsection{Evaluation of Rendering}
Table \ref{render} presents a comparison of rendering results from all reproducible open-source algorithms implemented on the same platform. 3DGS and NeRF++ do not perform pose estimation and utilize COLMAP to obtain the initial inputs. SplaTAM, MonoGS and Gaussian-SLAM rely on loss functions to optimize pose, so we replicated the rendering results twice: once using the ground truth poses and once using the poses estimated by themselves.

Fig. \ref{Comparison of Render} presents the qualitative results of image rendering, the rendering quality of 3DGS is affected by the lack of depth priors and the limited perspectives obtained during the SLAM motion process. The images rendered by LiV-GS exhibit remarkable clarity, capturing nuanced details such as the architectural structure and outlines of buildings, ground textures, foliage details of trees and shrubs, and even reflections on car hoods. The fidelity in rendering not only demonstrates the effectiveness of LiV-GS but also highlights its capability in preserving intricate visual details.

\subsection{Evaluation between performance and runtime}
To further estimate the tradeoff between efficiency and performance of LiV-GS, we compared the relationship between runtime and previously evaluated performance metrics. The ATE RMSE is directly used to measure localization accuracy and the image quality  of each algorithm is  calculated by normalizing the composite score:
\[
\text{Image Quality} = \text{SSIM} + \text{PSNR}/30 + (1 - \text{LPIPS})
\]

As shown in Figs. \ref{Image quality} and \ref{absolute error},  LiV-GS simultaneously reaches state-of-the-art performance in both accuracy and rendering quality with a running speed of 7.98 FPS, showing its potential for real-time SLAM application as the LiDAR sampling rate is 10 Hz.

As to the running time of each module,  the front-end tracking and keyframe selection module operates at 0.07 ms per operation on average. With the back-end configured with five-keyframe optimization at a time, the average runtime of the pose optimization and map update modules are 0.04 ms and  0.09 ms, respectively. It is worth noting that our LiV-GS system employs an asynchronous communication mechanism, so the operating time of the entire system does not equal the sum of the running time of individual modules.

\subsection{Qualitative Analysis of Geometry Accuracy}
To assess the spatial geometric distribution of Gaussian map constructed by LiV-GS, we conducted cross-modal mmWave radar localization on the Gaussian map. Unlike LiDAR, the point clouds of mm-Wave radar are much sparser and have lower resolution and higher noises. 
Robust cross-modal radar localization is possible only if the spatial geometry of the Gaussian map generated by LiDAR and images is sufficiently accurate and reliable. 

For the sequence cp from NTU4DRadLM dataset, we first constructed the Gaussian map based on camera and LiDAR data using our LiV-GS, then relocalized radar data on the Gaussian map using the HDL-localization algorithm \cite{koide2019hdl}.  Fig. 8 highlights that even with cross-modal radar data, accurate localization is consistently achieved using Gaussian maps. It confirms that the Gaussian map of LiV-GS provides accurate geometric structure information, showing its great potential for other downstream tasks such as all-weather outdoor localization and navigation.

\begin{figure}[t]
    \centering
    \includegraphics[width=\linewidth]{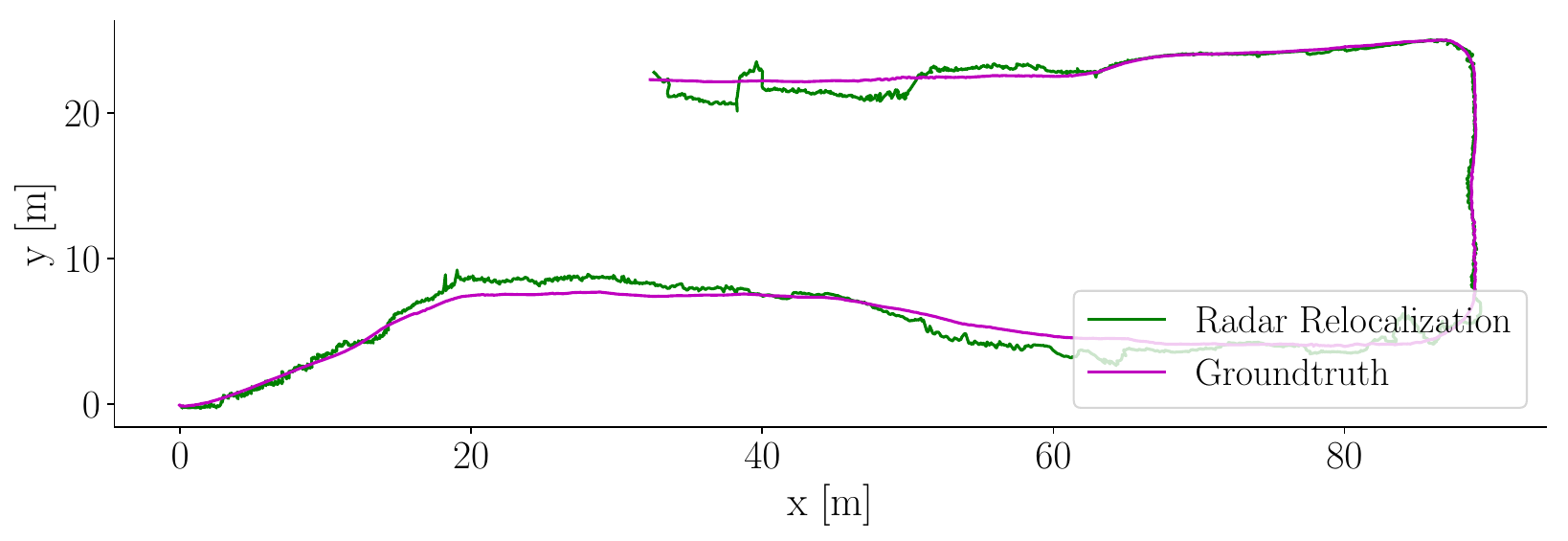}
    \caption{\textbf{Visualization of cross-modal mmWave radar localization trajectory.} }
    \label{relocalization}
\vspace{-5mm}
\end{figure}

\subsection{Close-loop Validation}
In this subsection, we verify the performance of LiV-GS in closed-loop sequences.  We intercepted a closed-loop sequence of the hku$\_$park$\_$00  data from the handheld scanning dataset provided by R$^{3}$Live \cite{lin2022r3live}, which includes $1280 \times 1024$ images at 30Hz and Livox-Avia LiDAR point clouds at 10Hz.

The quantitative results are shown in Table \ref{table:tracking_accuracy} and \ref{table:rendering}. In this looped sequence, our LiV-GS still performs well but its performance falls behind some other algorithms occasionally. One reason for that is LiV-GS lacks a loop closure detection module to optimize the accumulated drift. On the other hand, Since LiV-GS does not utilize spherical harmonics to represent colors. Observing different colors of the same scene from varying viewpoints at loop closure positions could result in a substantial overlap of Gaussian ellipsoids, which might lead to performance degradation. How to address the issues of loop closure and color refinement will be the focus of our future research.

\begin{table}[t]
\centering
\begin{minipage}{0.35\textwidth}
\centering
\caption{\footnotesize{Quantitative Analysis for Tracking}}
\label{table:tracking_accuracy}
\resizebox{0.9\textwidth}{!}{
\begin{tabular}{lccc}
\toprule
% \multicolumn{4}{c}{\textbf{HKU Park}} \\
% \cmidrule(lr){2-4}
\textbf{Methods}  & $t_{rel}\downarrow$ & $r_{rel}\downarrow$ & $t_{abs}\downarrow$ \\
\midrule
NeRF-LOAM                   & 0.518 & 0.636  & 1.617 \\
HDL-graph-SLAM              & \underline{0.497} & 0.801 & 0.712 \\
ORB-SLAM3                   & 0.446 & 0.712 & \textbf{0.457} \\
SplaTAM                     & - & - & - \\
MonoGS                      & 0.535 & 0.932 & 1.963 \\
Gaussian-SLAM               & 1.456 & \underline{0.562} & 1.444 \\
GS-ICP-SLAM                 & - & - & - \\
Ours                        & \textbf{0.436} & \textbf{0.546} & \underline{0.705} \\
\bottomrule
\end{tabular}
}
\end{minipage}%
\hfill
\begin{minipage}{0.35\textwidth}
\centering
\caption{\footnotesize{Quantitative Analysis for Rendering}}
\label{table:rendering}
\resizebox{0.9\textwidth}{!}{
\begin{tabular}{lccc}
\toprule
% \multicolumn{4}{c}{\textbf{HKU Park}} \\
% \cmidrule(lr){2-4}
\textbf{Methods} & \textbf{SSIM$\uparrow$} & \textbf{PSNR$\uparrow$} & \textbf{LPIPS$\downarrow$} \\
\midrule
3DGS                     & \textbf{0.868} & \textbf{25.776} & \textbf{0.247} \\
GS-ICP-SLAM               & 0.446 & 11.949 & 0.813 \\
NeRF++                    & 0.581 & 12.256 & 0.791 \\
SplaTAM              & 0.235 & 10.501 & 0921 \\
MonoGS               & 0.533 & 15.928 & 0.694 \\
Gaussian-SLAM        & 0.399 & 10.523 & 0.928 \\
Ours                 & \underline{0.565} & \underline{17.316} & \underline{0.568}\\
\bottomrule
\end{tabular}
}
\end{minipage}
\vspace{-5mm}
\end{table}

% \begin{figure}[t]
%     \centering
%     \includegraphics[width=\linewidth]{figures/pdf/park.pdf}
%     \caption{\textbf{Comparison of Rendering Results in HKU park.}}
%     \label{Comparison in park}
% \end{figure}

\section{Conclusion}
We propose a novel outdoor SLAM system leveraging 3D gaussian as spatial representation, integrating LiDAR-camera data. The system tightly integrates LiDAR point clouds with Gaussian maps via covariance for tracking and uses visual information to optimize the global Gaussian distribution. Conditional Gaussian constraints guide Gaussian ellipsoid splitting, especially for Gaussian ellipsoids lacking geometric constraints as LiDAR depth is unavailable. This method exploits visual texture continuity and LiDAR reliability to construct ellipsoids with consistent depth constraints.

Experiments validate that our system achieves robust, precise localization and can render clear scene imagery. To our knowledge, this is the first successful implementation of cross-modal radar-LiDAR localization using  3D Gaussian maps in outdoor environments, marking a significant potential application of outdoor 3DGS SLAM.

\bibliographystyle{IEEEtran}
\bibliography{IEEEabrv , reference}

\begin{thebibliography}{10}
\providecommand{\url}[1]{#1}
\csname url@samestyle\endcsname
\providecommand{\newblock}{\relax}
\providecommand{\bibinfo}[2]{#2}
\providecommand{\BIBentrySTDinterwordspacing}{\spaceskip=0pt\relax}
\providecommand{\BIBentryALTinterwordstretchfactor}{4}
\providecommand{\BIBentryALTinterwordspacing}{\spaceskip=\fontdimen2\font plus
\BIBentryALTinterwordstretchfactor\fontdimen3\font minus \fontdimen4\font\relax}
\providecommand{\BIBforeignlanguage}[2]{{%
\expandafter\ifx\csname l@#1\endcsname\relax
\typeout{** WARNING: IEEEtran.bst: No hyphenation pattern has been}%
\typeout{** loaded for the language `#1'. Using the pattern for}%
\typeout{** the default language instead.}%
\else
\language=\csname l@#1\endcsname
\fi
#2}}
\providecommand{\BIBdecl}{\relax}
\BIBdecl

\bibitem{mildenhall2021nerf}
B.~Mildenhall, P.~P. Srinivasan, M.~Tancik, J.~T. Barron, R.~Ramamoorthi, and R.~Ng, ``Ne{RF}: Representing {S}cenes as {N}eural {R}adiance {F}ields for view synthesis,'' \emph{Communications of the ACM}, vol.~65, no.~1, pp. 99--106, 2021.

\bibitem{kerbl20233dgs}
B.~Kerbl, G.~Kopanas, T.~Leimk{\"u}hler, and G.~Drettakis, ``3{D} {G}aussian {S}platting for {R}eal-{T}ime {R}adiance {F}ield {R}endering.'' \emph{ACM Trans. Graph.}, vol.~42, no.~4, pp. 139--1, 2023.

\bibitem{deng2023nerfloam}
J.~Deng, Q.~Wu, X.~Chen, S.~Xia, Z.~Sun, G.~Liu, W.~Yu, and L.~Pei, ``Ne{RF}-{L}{O}{A}{M}: {N}eural {I}mplicit {R}epresentation for {L}arge-scale incremental {L}i{D}{A}{R} odometry and mapping,'' in \emph{Proceedings of the IEEE/CVF International Conference on Computer Vision}, 2023, pp. 8218--8227.

\bibitem{ha2024rgbdgsicpslam}
S.~Ha, J.~Yeon, and H.~Yu, ``{R}{G}{B}{D} {G}s-icp {S}{L}{A}{M},'' \emph{arXiv preprint arXiv:2403.12550}, 2024.

\bibitem{tao2024silvr}
Y.~Tao, Y.~Bhalgat, L.~F.~T. Fu, M.~Mattamala, N.~Chebrolu, and M.~Fallon, ``Si{L}{V}{R}: Scalable {L}i{D}{A}{R}-{V}isual {R}econstruction with {N}eural {R}adiance {F}ields for robotic inspection,'' \emph{arXiv preprint arXiv:2403.06877}, 2024.

\bibitem{yan2024gsslam}
C.~Yan, D.~Qu, D.~Xu, B.~Zhao, Z.~Wang, D.~Wang, and X.~Li, ``G{S}-{S}{L}{A}{M}: Dense visual {S}{L}{A}{M} with 3{D} {G}aussian {S}platting,'' in \emph{Proceedings of the IEEE/CVF Conference on Computer Vision and Pattern Recognition}, 2024, pp. 19\,595--19\,604.

\bibitem{sun2024mm3dgs}
L.~C. Sun, N.~P. Bhatt, J.~C. Liu, Z.~Fan, Z.~Wang, T.~E. Humphreys, and U.~Topcu, ``M{M}3{D}{G}{S} {S}{L}{A}{M}: {M}ulti-{M}odal 3{D} {G}aussian {S}platting for {S}{L}{A}{M} {U}sing {V}ision, {D}epth, and {I}nertial {M}easurements,'' \emph{arXiv preprint arXiv:2404.00923}, 2024.

\bibitem{islam2024mvslam}
Q.~U. Islam, H.~Ibrahim, P.~K. Chin, K.~Lim, and M.~Z. Abdullah, ``{M}{V}{S}-{S}{L}{A}{M}: {E}nhanced {M}ultiview geometry for improved {S}emantic {R}{G}{B}{D} {S}{L}{A}{M} in dynamic environment,'' \emph{Journal of Field Robotics}, vol.~41, no.~1, pp. 109--130, 2024.

\bibitem{kerbl2024hierarchical}
B.~Kerbl, A.~Meuleman, G.~Kopanas, M.~Wimmer, A.~Lanvin, and G.~Drettakis, ``A {H}ierarchical 3{D} {G}aussian {R}epresentation for {R}eal-{T}ime rendering of very large datasets,'' \emph{arXiv preprint arXiv:2406.12080}, 2024.

\bibitem{chen2024dogaussian}
Y.~Chen and G.~H. Lee, ``Do{G}aussian: {D}istributed-{O}riented {G}aussian {S}platting for {L}arge-{S}cale 3{D} {R}econstruction {V}ia {G}aussian {C}onsensus,'' \emph{arXiv preprint arXiv:2405.13943}, 2024.

\bibitem{xie2024gaussiancity}
H.~Xie, Z.~Chen, F.~Hong, and Z.~Liu, ``Gaussian{C}ity: {G}enerative {G}aussian {S}platting for {U}nbounded 3{D} {C}ity {G}eneration,'' \emph{arXiv preprint arXiv:2406.06526}, 2024.

\bibitem{ren2024octree-gs}
K.~Ren, L.~Jiang, T.~Lu, M.~Yu, L.~Xu, Z.~Ni, and B.~Dai, ``Octree-{G}{S}: {T}owards consistent {R}eal-{T}ime rendering with {L}{o}{D}-structured 3{D} {G}aussians,'' \emph{arXiv preprint arXiv:2403.17898}, 2024.

\bibitem{liu2024efficientgs}
W.~Liu, T.~Guan, B.~Zhu, L.~Ju, Z.~Song, D.~Li, Y.~Wang, and W.~Yang, ``Efficient{G}{S}: {S}treamlining {G}aussian {S}platting for large-scale high-resolution scene representation,'' \emph{arXiv preprint arXiv:2404.12777}, 2024.

\bibitem{cheng2024gaussianpro}
K.~Cheng, X.~Long, K.~Yang, Y.~Yao, W.~Yin, Y.~Ma, W.~Wang, and X.~Chen, ``Gaussian{P}ro: 3{D} {G}aussian {S}platting with progressive propagation,'' in \emph{Forty-first International Conference on Machine Learning}, 2024.

\bibitem{zhou2024drivinggaussian}
X.~Zhou, Z.~Lin, X.~Shan, Y.~Wang, D.~Sun, and M.-H. Yang, ``Driving{G}aussian: {C}omposite {G}aussian {S}platting for surrounding dynamic autonomous driving scenes,'' in \emph{Proceedings of the IEEE/CVF Conference on Computer Vision and Pattern Recognition}, 2024, pp. 21\,634--21\,643.

\bibitem{jiang20243dgsReloc}
P.~Jiang, G.~Pandey, and S.~Saripalli, ``3{D}{G}{S}-{R}eloc: 3{D} {G}aussian {S}platting for map representation and visual relocalization,'' \emph{arXiv preprint arXiv:2403.11367}, 2024.

\bibitem{lang2024gaussianLIC}
X.~Lang, L.~Li, H.~Zhang, F.~Xiong, M.~Xu, Y.~Liu, X.~Zuo, and J.~Lv, ``Gaussian-{L}{I}{C}: Photo-realistic {L}i{D}{A}{R}-inertial-camera {S}{L}{A}{M} with 3{D} {G}aussian {S}platting,'' \emph{arXiv preprint arXiv:2404.06926}, 2024.

\bibitem{hong2024livGaussianmap}
S.~Hong, J.~He, X.~Zheng, C.~Zheng, and S.~Shen, ``L{I}{V}-{G}auss{M}ap: {L}i{D}{A}{R}-{I}nertial-{V}isual {F}usion for {R}eal-{T}ime 3{D} {R}adiance {F}ield map rendering,'' \emph{IEEE Robotics and Automation Letters}, 2024.

\bibitem{zhao2024tclcgs}
C.~Zhao, S.~Sun, R.~Wang, Y.~Guo, J.-J. Wan, Z.~Huang, X.~Huang, Y.~V. Chen, and L.~Ren, ``T{C}{L}{C}-{G}{S}: Tightly coupled {L}i{D}{A}{R}-camera {G}aussian {S}platting for surrounding autonomous driving scenes,'' \emph{arXiv preprint arXiv:2404.02410}, 2024.

\bibitem{cui2024letsgo}
J.~Cui, J.~Cao, Y.~Zhong, L.~Wang, F.~Zhao, P.~Wang, Y.~Chen, Z.~He, L.~Xu, Y.~Shi \emph{et~al.}, ``Lets{G}o: Large-scale garage modeling and rendering via {L}i{D}{A}{R}-assisted gaussian primitives,'' \emph{arXiv preprint arXiv:2404.09748}, 2024.

\bibitem{wu2024mmgaussian}
C.~Wu, Y.~Duan, X.~Zhang, Y.~Sheng, J.~Ji, and Y.~Zhang, ``M{M}-{G}aussian: 3{D} {G}aussian-based multi-modal fusion for localization and reconstruction in unbounded scenes,'' \emph{arXiv preprint arXiv:2404.04026}, 2024.

\bibitem{matsuki2024gaussianslams}
H.~Matsuki, R.~Murai, P.~H. Kelly, and A.~J. Davison, ``Gaussian {S}platting {S}{L}{A}{M},'' in \emph{Proceedings of the IEEE/CVF Conference on Computer Vision and Pattern Recognition}, 2024, pp. 18\,039--18\,048.

\bibitem{guedon2024sugar}
A.~Gu{\'e}don and V.~Lepetit, ``Su{G}a{R}: {S}urface-aligned {G}aussian {S}platting for efficient 3d mesh reconstruction and high-quality mesh rendering,'' in \emph{Proceedings of the IEEE/CVF Conference on Computer Vision and Pattern Recognition}, 2024, pp. 5354--5363.

\bibitem{behley2018SUMA}
J.~Behley and C.~Stachniss, ``Efficient {S}urfel-based {S}{L}{A}{M} using 3d {L}aser range data in urban environments.'' in \emph{Robotics: science and systems}, vol. 2018, 2018, p.~59.

\bibitem{Zhang18irosrpg}
Z.~Zhang and D.~Scaramuzza, ``A {T}utorial on {Q}uantitative {T}rajectory {E}valuation for {V}isual(-{I}nertial) {O}dometry,'' in \emph{IEEE/RSJ Int. Conf. Intell. Robot. Syst. (IROS)}, 2018.

\bibitem{lin2022r3live}
J.~Lin and F.~Zhang, ``R3{L}{I}{V}{E}: A {R}obust, {R}eal-{T}ime, {R}{G}{B}-colored, {L}i{D}{A}{R}-{I}nertial-{V}isual tightly-coupled state {E}stimation and mapping package,'' in \emph{2022 International Conference on Robotics and Automation (ICRA)}.\hskip 1em plus 0.5em minus 0.4em\relax IEEE, 2022, pp. 10\,672--10\,678.

\bibitem{koide2019hdl}
K.~Koide, J.~Miura, and E.~Menegatti, ``A {P}ortable {T}hree-{D}imensional {L}i{D}{A}{R}-based system for long-term and wide-area people behavior measurement,'' \emph{International Journal of Advanced Robotic Systems}, vol.~16, no.~2, p. 1729881419841532, 2019.

\bibitem{campos2021orbslam3}
C.~Campos, R.~Elvira, J.~J.~G. Rodr{\'\i}guez, J.~M. Montiel, and J.~D. Tard{\'o}s, ``O{R}{B}-{S}{L}{A}{M}3: {A}n accurate open-source library for visual, visual--inertial, and multimap {S}{L}{A}{M},'' \emph{IEEE Transactions on Robotics}, vol.~37, no.~6, pp. 1874--1890, 2021.

\bibitem{yugay2023gaussianslam}
V.~Yugay, Y.~Li, T.~Gevers, and M.~R. Oswald, ``Gaussian-{S}{L}{A}{M}: {P}hoto-realistic dense {S}{L}{A}{M} with {G}aussian {S}platting,'' \emph{arXiv preprint arXiv:2312.10070}, 2023.

\bibitem{keetha2024splatam}
N.~Keetha, J.~Karhade, K.~M. Jatavallabhula, G.~Yang, S.~Scherer, D.~Ramanan, and J.~Luiten, ``Spla{T}{A}{M}: {S}plat {T}rack \& map 3{D} {G}aussians for {D}ense {R}{G}{B}-{D} {S}{L}{A}{M},'' in \emph{Proceedings of the IEEE/CVF Conference on Computer Vision and Pattern Recognition}, 2024, pp. 21\,357--21\,366.

\bibitem{zhang2020nerf++}
K.~Zhang, G.~Riegler, N.~Snavely, and V.~Koltun, ``Ne{R}{F}++: {A}nalyzing and improving {N}eural {R}adiance {F}ields,'' \emph{arXiv preprint arXiv:2010.07492}, 2020.

\end{thebibliography}
\end{document}